\documentclass[10pt,journal,compsoc]{IEEEtran}
\ifCLASSOPTIONcompsoc
  \usepackage[nocompress]{cite}
\else
  \usepackage{cite}
\fi
\ifCLASSINFOpdf
\else
\fi

\usepackage{subcaption}
\usepackage{graphicx}
\usepackage{url} 
\usepackage{amsmath}
\usepackage{makecell}
\usepackage{multirow}
\usepackage{booktabs}
\usepackage{color}
\usepackage{float}
\usepackage{amssymb}
\usepackage{mathrsfs}
\usepackage{comment}
\usepackage{color,array}

\makeatletter

\def\zapcolorreset{\let\reset@color\relax\ignorespaces}
\def\colorrows#1{\noalign{\aftergroup\zapcolorreset#1}\ignorespaces}

\edef\endtabular{\unexpanded\expandafter{\endtabular}\noexpand\reset@color} 

\makeatother

\newcommand\model{GAMCN}

\begin{document}
%
\title{A Graph and Attentive Multi-Path Convolutional Network for Traffic Prediction}
%
%
%
%

\author{Jianzhong~Qi,
        Zhuowei~Zhao, 
        Egemen~Tanin,
        Tingru~Cui,
        Neema~Nassir, 
        and Majid~Sarvi
\IEEEcompsocitemizethanks{\IEEEcompsocthanksitem J. Qi (Corresponding author), Z. Zhao, E. Tanin, and T. Cui are with the 
 School of Computing and Information Systems, The University of Melbourne, Victoria 3010 Australia.\protect\\
E-mail: \{jianzhong.qi, zhuowei.zhao, etanin, tingru.cui\}@unimelb.edu.au
}
\IEEEcompsocitemizethanks{\IEEEcompsocthanksitem N. Nassir and M. Sarvi are with the 
Department of Infrastructure Engineering, The University of Melbourne, Victoria 3010 Australia.\protect\\
E-mail: \{neema.nassir, majid.sarvi\}@unimelb.edu.au
}
\thanks{Manuscript received June 07, 2021; revised March 25, 2022.}
}

%
%

\markboth{IEEE Transactions on Knowledge and Data Engineering,~Vol.~XX, No.~X, MAY~2022}%
{Qi \MakeLowercase{\textit{et al.}}: A Graph and Attentive Multi-Path Convolutional Network for Traffic Prediction}
%



\IEEEtitleabstractindextext{%
\begin{abstract}
Traffic prediction is an important and yet highly challenging problem due to the complexity and constantly changing nature of traffic systems. To address the challenges, we propose a \emph{graph and attentive multi-path convolutional network} (\model) model to predict traffic conditions such as traffic speed across a given road network into the future. Our model focuses on the spatial and temporal factors that impact traffic conditions. To model the spatial factors, we propose a variant of the \emph{graph convolutional network}~(GCN) named \emph{LPGCN} to embed road network graph vertices into a latent space, where vertices with correlated traffic conditions are close to each other. To model the temporal factors, we use a  multi-path \emph{convolutional neural network} (CNN) to learn the joint impact of different combinations of past traffic conditions on the future traffic conditions. Such a joint  impact is further modulated by an \emph{attention} generated from an embedding of the prediction time, which encodes the periodic patterns of traffic conditions. We evaluate our model on real-world road networks and traffic data. The experimental results show that our  model outperforms state-of-art traffic prediction models by up to 18.9\% in terms of prediction errors and 23.4\% in terms of prediction efficiency.
\end{abstract}

} 

\maketitle

\IEEEdisplaynontitleabstractindextext

%
\IEEEpeerreviewmaketitle

\IEEEraisesectionheading{\section{Introduction}\label{sec:introduction}}

%
%
%
%
\IEEEPARstart{T}{raffic} prediction aims to predict the traffic conditions (e.g., traffic  speed and volume) into the future given a road network and historical traffic condition observations (e.g., recorded via sensors). It plays an important role in many real-world applications. For example, traffic prediction can help offer more accurate travel time predictions of different routes, such that travelers can make effective informed decisions in route planning. As another example, accurate traffic prediction can help traffic management systems make effective informed decisions (e.g., to open or close certain lanes) to reduce traffic congestion~\cite{zheng2020gman}. 

\begin{figure}[ht]
\vspace{-5mm}
\begin{subfigure}{.49\linewidth}
  \includegraphics[width=1.10\linewidth]{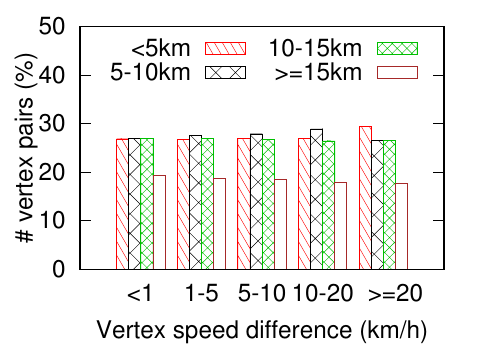}  
  \caption{Vertex distance distribution vs. speed differences}
  \label{fig:spatial_distribution_pems04}
\end{subfigure}
\hspace{1mm}
\begin{subfigure}{.49\linewidth}
\hspace{-5mm}
  \includegraphics[width=1.10\linewidth]{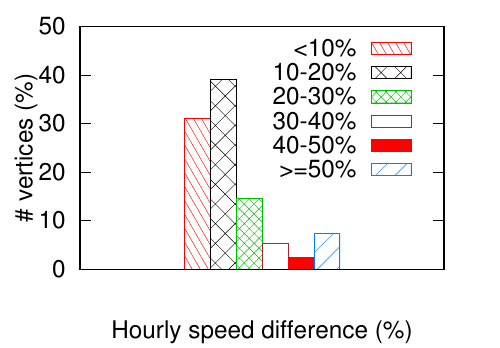}  
  \caption{Hourly speed change distribution}
  \label{fig:temporal_distribution_pems04}
\end{subfigure}
\vspace{-2mm}
\caption{Spatial and temporal correlations of traffic speed}\label{fig:motivation}
\vspace{-3mm}
\end{figure}

Traffic prediction is challenging due to the highly complex nature of traffic systems and many impacting factors. In this paper, we focus on the spatial and temporal factors. From the spatial perspective, intuitively, points nearby in a road network are more likely to share similar traffic patterns. However, this is not always the case. 
To showcase this observation, we collect the (District 4) traffic speed data recorded by 400 sensors between 1st of December 2020 and 31st of January 2021 at 5-minute interval from the Caltrans Performance Measurement System (PeMS, \url{https://pems.dot.ca.gov}). 
There are a total of  17,856 time intervals (i.e., 12 intervals per hour $\times$ 24 hours per day $\times$ 62 days. For every time interval, we calculate the speed difference between every pair of vertices. There are 18.18\%, 49.54\%, 18.72\%, 7.01\%, and 6.55\% of the vertex pairs with a speed difference of less than 1 km/h,  1 to 5 km/h, 5 to 10 km/h,  10 to 20 km/h, and greater than or equal to 20 km/h, respectively. 
For the vertex pairs in each speed difference category, we further calculate the (road network) distance between the two vertices in each pair. 
Fig.~\ref{fig:spatial_distribution_pems04} reports the percentage of vertex pairs of different distances. We see that, vertex pairs with a small speed difference are not necessarily close to each other, e.g., 26.82\%, 26.93\%, 26.94\%, and 19.31\% of the vertex pairs with a speed difference of less than 1 km/h have a distance of  less than 5 km,  5 to 10 km,   10 to 15 km, and greater than or equal to 15 km, respectively. Note that the maximum pairwise vertex distance is 26.6 km. 
The vertices with a small traffic speed difference (i.e., $< 1$~km/h) but a large distance (e.g., $\ge 10$ km, representing 18.18\% $\times$ (26.94\%+19.31\%) = 8.41\% of all vertex pairs) may be missed by existing models~\cite{geng2019spatiotemporal, li2017diffusion, yu2017spatio} that use \emph{graph convolutional networks} (GCNs) directly, which focus on nearby vertices in a road network graph. 

From the temporal perspective, traffic conditions at different points in a road network  may  change differently with time. 
In Fig.~\ref{fig:temporal_distribution_pems04}, we further report the percentage of vertices  with different speed changes accumulated in an hour: 31.06\%, 38.99\%, and 7.36\% of the vertices have accumulated less than 10\%, 10\% to 20\%, and more than 
50\% changes in the speed, respectively.
To uncover such different changing patterns, it is critical to learn the joint impact of past traffic conditions on the  future traffic conditions for each point. \emph{Recurrent neural networks} (RNN) have been used to model such temporal correlations~\cite{geng2019spatiotemporal, li2017diffusion, li2019learning, pan2019urban}, which have limitations in the computation efficiency and may suffer when being applied on longer input sequences.  \emph{Convolutional neural networks} (CNN)~\cite{yu2017spatio,wu2019graph} have also been used, for their higher efficiency and capability in handling longer sequences without forgetting the early  signals. The state-of-the-art model, the \emph{graph multi-attention network} (GMAN)~\cite{zheng2020gman}, models the spatio-temporal correlations with \emph{attention networks}. This model, as will be shown in our experiments, suffers in modeling the spatial correlations due to its simple attention-based structure, and hence may not produce the most accurate predictions for the immediate future.  

To better model the spatial and temporal correlations for higher prediction accuracy, we propose a  \emph{graph and attentive multi-path convolutional network} (\model) model that takes the traffic conditions at points in a road network (which are modeled as vertices in a graph) in the past $p$ time points as the input and predicts the traffic conditions for $q$ time points into the future, where $p$ and $q$ are system parameters determined by application requirements. 

 \model\  learns the spatial correlations via a GCN variant named \emph{LPGCN} that we design to model the latent connectivity among the vertices in a road network graph 
 using a  learnable \emph{point-wise mutual information} (PMI) matrix. 
 This matrix  is learned together with the vertex embeddings.  
 When it is learned, a larger matrix element indicates a pair of vertices with more similar traffic conditions, i.e., a stronger connection in a latent (spatial correlation) space. The matrix can be used together with a graph adjacency matrix for  spatial correlation modeling, or on its own to model the (latent) vertex connectivity when the adjacency matrix is unavailable. This makes our model applicable to scenarios where adjacency matrices are unavailable. 
 
 \model\ learns the temporal correlations via  multiple 1-dimensional (1-D) CNNs  (i.e., a \emph{multi-path CNN}) to learn the joint impact of past traffic observations. Such a joint impact is further modulated by an attention generated from the \emph{prediction time}, i.e., the time for which the traffic conditions are to be predicted. To encode the periodicity of traffic conditions, we use one-hot embeddings to represent  the time of day and the day of week of the prediction time.  
 
 We use a multi-path CNN because the traffic condition to be predicted may be impacted by traffic observations from different combinations of past time points. For example, the average traffic speed of the \emph{past hour} may be a robust estimate for the traffic speed in the next 15 minutes in non-peak hours, while the average traffic speed of the \emph{past 5 minutes} may be a more accurate estimate when a peak hour has just started. Our multi-path CNN has a different kernel size for each path. Thus, each path learns the impact of a different past time combination. The importance of the output from  different CNN paths may vary at different prediction times (e.g., peak hour vs. non-peak hour). Our attention mechanism helps learn such varying importance weights given different input prediction times. 
Our model design differs from existing models that use CNNs with attention networks: (1) The \emph{AttConv} model~\cite{yin-schutze-2018-attentive} takes the output of an attention computation to modify the input of a CNN, while we compute attention on the output of multiple CNN paths to aggregate them.  
(2)~The \emph{SKNets} model~\cite{DBLP:conf/cvpr/LiW0019} computes attention weights from intermediate output of a multi-path CNN, while the attention in our model comes from outside the multi-path CNN (i.e., the time information).  
(3)~The  \emph{attentive group equivariant convolutional network}~\cite{pmlr-v119-romero20a} uses two CNN groups on the same input, where one group yields the attention and the other yields the signals, while all our CNN paths are used to yield the signals,  and the attention comes from the time information as mentioned above.

In summary, we make the following contributions:

(1) We propose a novel deep neural network model named \model\ 
    to predict future traffic conditions (e.g., for an hour ahead of time).

(2) To model the (spatial) correlations of traffic conditions at different vertices in a road network graph, we propose a GCN-based model named LPGCN that learns a transition matrix to encode the latent traffic pattern correlations among the vertices. This learned matrix can be used either together with a graph adjacency matrix or to replace the adjacency matrix, which enables our model to be used for applications even without knowing the adjacency matrices.
    
(3) To model the (temporal) correlations of traffic conditions at different time points, we propose to learn the joint impact of traffic conditions observed at different combinations of past time points, using a multi-path CNN. Each CNN path focuses on a different combination. The learned joint impact is further modulated by an attention mechanism using an embedding of the traffic prediction time, to reflect the periodic patterns of traffic conditions. 

(4) We run experiments on real-world road networks and traffic data. The results show that: (1) \model\ outperforms the state-of-art traffic prediction models by up to 18.9\% in prediction errors and 23.4\% in prediction efficiency; (2)  \model\ can be easily combined with a \emph{generative adversarial network} to make predictions that are not only highly accurate but also share a similar distribution with the ground truth. 
   
\section{Related Work}\label{sec:lit_review}

We review both traditional and recent deep learning-based methods for traffic prediction.

\subsection{Traditional Methods}
Traditional studies on traffic prediction use statistical or classic machine learning methods. 
For example, a series of studies~\cite{williams2003modeling,shekhar2007adaptive,li2012prediction} use \emph{autoregressive integrated moving average} (ARIMA) to model the temporal correlations among traffic conditions at different times. Williams and Hoel~\cite{williams2003modeling} first analyze the performance of seasonal ARIMA on traffic flow prediction. They compare ARIMA with \emph{random walk}, \emph{historical average}, and \emph{deviation from historical average}  and show that seasonal ARIMA yields the best performance. 
Shekhar and Williams~\cite{shekhar2007adaptive} propose an adaptive ARIMA model with dynamical filters that update the parameters as new data is arriving, for short-term traffic flow prediction. Li et al.~\cite{li2012prediction} improve ARIMA by considering both the nearest historical data and the periodic data (such as data from a week ago) to predict taxi demands. 
Such methods ignore  the spatial correlations, and they may suffer in the prediction accuracy for complex road networks where the traffic conditions at different places may be correlated. 

Classic machine learning methods such as regression models~\cite{li2018general} have been used to predict traffic based on human engineered features, such as orders placed by passengers, drivers' trajectories, points of interest (POI), and weather. 
Such methods require extensive human efforts in feature engineering and training data preparation. 
\emph{Gaussian process} models have also been used~\cite{diao2018hybrid, salinas2019high}, which consider both the spatial and the temporal  correlations. They yield accurate traffic predictions, but their scalability is an issue due to the high computational costs of Gaussian process.

\subsection{Deep Learning-Based Methods}
Recently, deep learning models have been introduced to traffic prediction and have shown promising results. A variety of deep neural networks are used to model the spatial and temporal correlations of traffic conditions.

\textbf{Spatial correlation modeling.} 
\emph{Convolutional neural networks} (CNN) are often used to model the spatial features for  traffic prediction. For example, 2-dimensional  (2-D) CNN models are proposed to  learn spatial features in an Euclidean space~\cite{ma2017learning, zhang2016dnn, zhang2017deep}. The intuition is that traffic conditions at nearby locations are correlated, while CNN models are effective to model such correlations. 
Ma et al.~\cite{ma2017learning}  convert traffic speed on different roads at different time slots into an ``image" (i.e., a matrix) and applied a CNN on the matrix for traffic speed prediction. 
Zhang et al.~\cite{zhang2016dnn} apply three CNNs to learn from traffic data collected at  most recent times, less recent times, and distant times, respectively. They then fuse the output of  these CNNs to produce the final traffic prediction. 
Zhang et al.~\cite{zhang2017deep} follow a similar idea and add extra features (weather and holidays) into the fusion. Wu and Tan~\cite{wu2016short} combine a 1-D CNN for spatial correlation learning with  \emph{long short-term memory} (LSTM) networks for temporal correlation learning. 

Later studies consider the constraints on traffic movement patterns imposed by the underlying road networks. For example, traffic conditions on both sides of a river without a bridge may vary substantially.  
To model such constraints, Lv et al.~\cite{DBLP:conf/ijcai/LvX0YZZ18} embed the road network topology into CNN (i.e., the \emph{look-up convolution}) by constructing filters based on road segment connectivity. \emph{Graph convolutional networks} (GCN) are often used~\cite{geng2019spatiotemporal,yu2017spatio,guo2019attention,DBLP:conf/nips/0001YL0020,DBLP:conf/aaai/SongLGW20,DBLP:conf/aaai/LiZ21}, as road networks can be easily represented by  graphs, and GCNs are highly effective for graph prediction problems. In GCNs, the feature signals (e.g., traffic speed values) are propagated among graph vertices based on their  adjacency matrix. In the simplest form, the features of a vertex $v$ only affect the immediate neighbors of $v$. 
The \emph{diffusion graph convolutional network} (DGCN)~\cite{li2017diffusion} replaces the adjacency matrix with a $k$-hop transition matrix in GCN. This model considers how the feature signals are propagated in $k$ hops, which can better model the impact  among vertices that are not directly connected.
Wu et al.~\cite{wu2019graph} propose a model named \emph{Graph WaveNet} that uses a learnable \emph{adaptive adjacency matrix} for GCNs  when the adjacency matrix is unknown. They initialize two matrices $\boldsymbol{E_1}$ and $\boldsymbol{E_2}$ with random values to represent the embeddings of vertices in a road network (which will be updated during training). The adaptive adjacency matrix is then defined as $\boldsymbol{\hat{A}} = softmax(relu(\boldsymbol{E_1} \cdot \boldsymbol{E_2}^T))$. Here, $relu(\cdot)$ denotes the \emph{ReLU} activation function, while $softmax(\cdot)$ denotes 
the \emph{softmax} function which re-scales the output of the calculation into range $(0, 1)$, such that the learned matrix $\boldsymbol{\hat{A}}$ can be seen as a normalized adjacency matrix.  In Graph WaveNet, 
the learnable adaptive adjacency matrix may be used on its own or together with DGCN if an adjacency matrix is given. 
A similar idea is used by \emph{AGCRN}~\cite{DBLP:conf/nips/0001YL0020} for adjacency matrix  learning, which further exploits this technique to improve GCN parameter learning efficiency. \emph{STSGCN}~\cite{DBLP:conf/aaai/SongLGW20} and  \emph{STFGNN}~\cite{DBLP:conf/aaai/LiZ21} construct spatio-temporal graphs using traffic observations of a few consecutive time points and use GCN on such graphs to learn spatial and temporal correlations.

Other recent studies~\cite{shi2020spatial,zheng2020gman} use the \emph{attention mechanism}~\cite{DBLP:conf/nips/VaswaniSPUJGKP17}. 
The intuition is to learn the traffic condition at a target vertex as  a weighted sum of those at the rest of the vertices. 
\emph{Spatial-temporal graph attention networks}  (ST-GAT)~\cite{shi2020spatial} generate attention with input traffic conditions alone, while the \emph{graph multi-attention network} (GMAN)~\cite{zheng2020gman} generates attention with input traffic conditions, road network embeddings, and time embeddings. The road network embeddings are learned in advance using \emph{node2vec}~\cite{grover2016node2vec}, while the time embeddings are one-hot  time vectors. GMAN yields the state-of-the-art traffic speed prediction accuracy. We compare with this model  empirically in Section~\ref{sec:exp}.

\textbf{Temporal correlation modeling.} CNN models have also been used to model the impact of nearby time points. For example, studies~\cite{yu2017spatio,wu2019graph} apply CNNs with a gated mechanism to learn the impact of the traffic conditions of past time points. They use two CNNs with different activation functions, \emph{sigmoid} and \emph{tanh}. The output of both CNNs is multiplied, such that the sigmoid CNN re-scales its output to range $(0,1)$ and works as a gate to control the strength of  the signals from the tanh CNN that go  through the gate. Unlike CNNs for spatial modeling, CNNs for temporal modeling are usually 1-D CNNs, i.e., only the time dimension is considered.

\emph{Recurrent neural networks} (RNN)~\cite{geng2019spatiotemporal}, \emph{gated recurrent unit} (GRU)~\cite{pan2019urban, li2017diffusion,DBLP:conf/nips/0001YL0020}, and LSTM~\cite{li2019learning} are also often used for temporal correlation modeling. These models are proposed for sequence learning problems, to which traffic prediction belongs. They take a sequence (e.g., traffic speed values for a few past time points) as the input and predict the observation of the next time point or  
next few time points. For example, Geng et al.~\cite{geng2019spatiotemporal} propose a \emph{contextual gated recurrent neural network} (CGRNN) that first weights the traffic data at a time point with a gated  mechanism based on the related vertices of a target vertex and then  applies an RNN to learn the temporal correlation. 
Li et al.~\cite{li2019learning} propose \emph{CE-LSTM} that learns from multi-source input data with an LSTM, which aims to retain more information from a longer input sequence than  RNNs do.    GRU has a simpler structure than LSTM and has also been shown to be effective for temporal correlation modeling~\cite{pan2019urban, li2017diffusion,DBLP:conf/nips/0001YL0020}.

\emph{Attention networks}~\cite{li2019forecaster, zheng2020gman} have also been used in temporal correlation modeling. Similar to that in spatial correlation modeling, the attention mechanism can learn the traffic condition at a target time point as a weighted sum of those at other (past) time points. 

RNN models, including its improved versions LSTM and GRU, have recursive structures which are expensive to train. In addition, sequence-to-sequence models based on them make predictions for one time point after another, which is prone to error propagation and long prediction times (when marking predictions for many time points ahead). In comparison, both CNN and attention networks have higher computation efficiency. Attention networks have shown both high efficiency and prediction accuracy~\cite{zheng2020gman}. They can also be used together with GCN~\cite{guo2019attention} or LSTM~\cite{shi2020spatial}. In our \model\ model, we use the attention mechanism with CNNs to model the temporal correlations.

A related problem is travel demand prediction (e.g., for taxi or ride-hailing services~\cite{DBLP:conf/kdd/WangYCW0019,DBLP:conf/icde/WangY0LWW021}), 
where recent travel demand records at a set of locations (or regions) are used to predict the travel demand for those locations in the near future. Spatio-temporal correlations also play an important role for this problem, and the techniques above such as graph neural networks, RNNs, and attention networks have been used for the problem. For future work, we plan to extend our \model\ model for  solving this problem. 

\textbf{Generative adversarial networks.} 
Recently, studies have applied \emph{generative adversarial networks} (GAN) for spatio-temporal data generation (e.g., to make predictions or fill in missing data). 
The basic idea is to use models similar to those described above as a generator to generate spatio-temporal data and a discriminator to reduce the difference in the distribution between the generated data and the real data. For example, Zhang et al.~\cite{zhang2021tits} propose a \emph{TrafficGAN} model for traffic prediction. Both their generator and discriminator use CNN and LSTM to model spatial and temporal correlations. 
Several studies~\cite{fernando2018gd, amirian2019social,sadeghian2019sophie} propose RNN based GAN structures for trajectory data prediction, while another series of studies~\cite{yang2019advanced,bojchevski2018netgan,lei2019gcn} propose GAN models for link predictions and graph representation learning. 
For example,  \emph{GCN-GAN}~\cite{lei2019gcn}  combines GCN and LSTM with GAN to predict dynamically weighted links in graphs. 
Overall, GAN-based models have shown promising results for spatio-temporal data generation. However, as a recent survey points out, the adaption of GANs for spatio-temporal data is still in its infancy~\cite{gao2020generative}. In Section~\ref{subsec:exp_gan}, we will extend our  model and study its applicability with GANs to further improve the quality of traffic predictions. 

\section{Proposed Model}\label{sec:model}
We consider a road network $G = \langle V, E \rangle$, where $V$ denotes a set of vertices (e.g., traffic sensors) and $E$ denotes a set of edges which represent the spatial connectivity (and distance) among the vertices.

Given historical traffic condition data (e.g., traffic speed values) at $V$ in the past $p$ time points, denoted as $\mathcal{X} =  (\boldsymbol{X_{t_1}}, \boldsymbol{X_{t_2}},  \ldots, \boldsymbol{X_{t_p}})$, we aim to predict the traffic conditions at $V$ for the next $q$ time points, denoted as $\boldsymbol{X_{t_p+1}}, \boldsymbol{X_{t_p+2}}, \ldots, \boldsymbol{X_{t_{p+q}}}$. Here, both $p$ and $q$ are system parameters. 
In $\boldsymbol{X_{t_i}} \in \mathbb{R}^{|V| \times c}$, each row has $c$ columns representing $c$ traffic conditions of interest such as speed, volume, etc. 
Each element can be used to represent a singular traffic condition value observed at a vertex at \emph{a single time point} $t_i$,  or an aggregate traffic condition value observed at a vertex over \emph{a period of time} $(t_{i-1}, t_i]$ (e.g., the total traffic volume or average traffic speed).  Thus, our proposed model can be applied for traffic prediction under both settings. 

\subsection{Model Overview}
To predict traffic conditions effectively and efficiently, we propose a model that learns the spatial and temporal correlations of traffic conditions at different places and time. Our model contains a variant of the GCN named LPGCN for spatial correlation modeling and a multi-path  CNN structure together with an attentive mechanism for temporal correlation modeling. Thus, we name our model the \emph{graph and attentive multi-path convolutional network} (\model). 

\begin{figure}
\hspace{-3mm}
\includegraphics[width=1.06\linewidth]{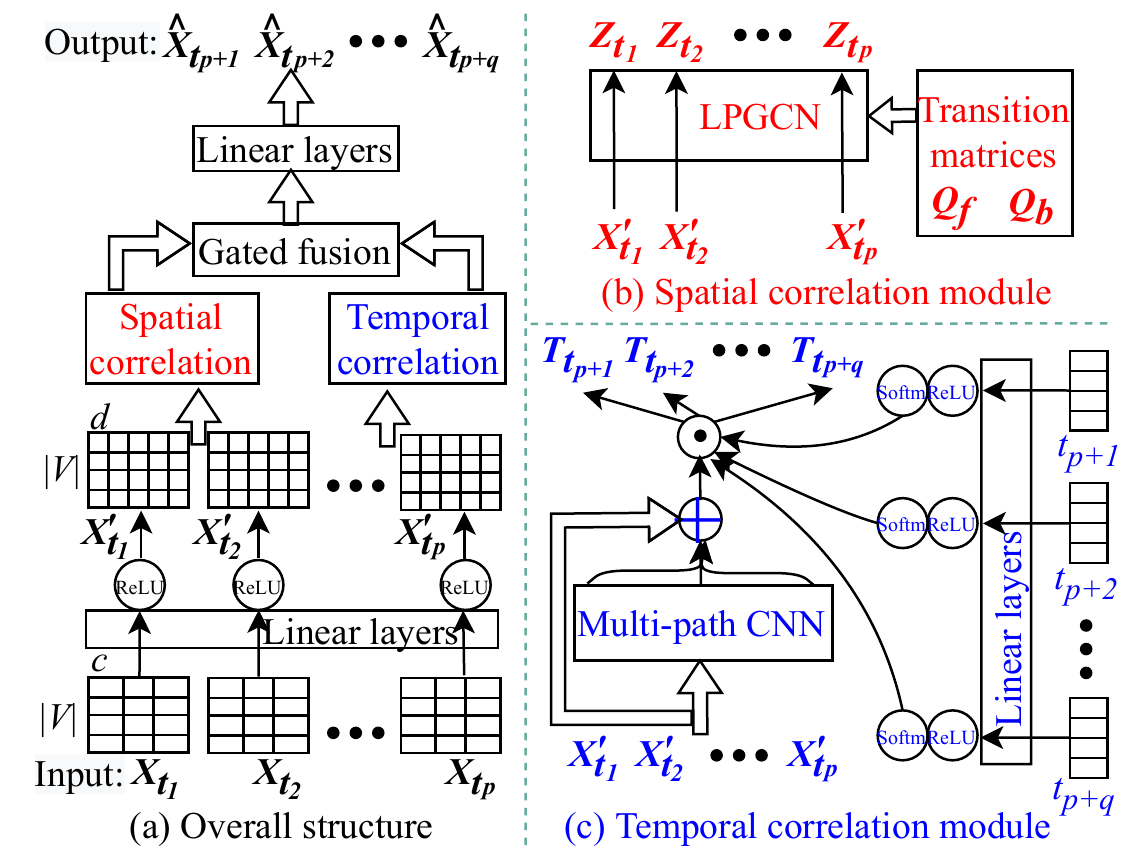}
\vspace{-6mm}
\caption{Model architecture of \model}
\label{fig:model} 
\vspace{-6mm}
\end{figure}

As shown in Fig.~\ref{fig:model}a,  \model\  takes a series of traffic data at all vertices of $G$ as the input. For the traffic data at each time point, $\boldsymbol{X_{t_i}}$, \model\ first uses a feedforward network with linear layers to map the traffic values to a higher dimensional latent space to enhance their representation capability (we use two layers of 10 and 100 nodes in our experiments). The output of the mapping, denoted as $\boldsymbol{X'_{t_i}} \in \mathbb{R}^{|V|\times d}$, is a $|V|\times d$ matrix where $d$ is the dimensionality of the latent space ($d = 100$ in our experiments).  
The output matrices of time $t_1$ to time $t_p$ go through two separate modules, i.e., the spatial correlation  and the temporal correlation modules, to learn the correlations of the traffic data:

(1) In the spatial correlation module (Fig.~\ref{fig:model}b): Each matrix $\boldsymbol{X'_{t_i}}$ is fed into a GCN variant named LPGCN that we propose to learn the spatial correlations  of the traffic conditions among the vertices (Section~\ref{subsec:spatial}).  

(2) In the temporal correlation module (Fig.~\ref{fig:model}c): The matrices of different time points together are fed into $p-1$ 1-dimensional CNNs (i.e., a multi-path CNN) to aggregate the historical observations at different time points. Each different CNN focuses on aggregating traffic data embeddings of a different time point combination. The output of all CNNs are concatenated to preserve the correlations of the traffic conditions of different time combinations. The concatenated output is then weighted by an attention generated based on the prediction time $t_{p+i}$ ($i \in [1, q]$) (Section~\ref{subsec:temporal}). 

The output of LPGCN and the temporal correlation module 
are fused with a gated fusion mechanism. The fused output is then fed into linear layers to generate the final traffic condition prediction (Section~\ref{subsec:gatefusion}). 

For model training, we use \emph{mean square error} (MSE) as our loss function $\mathcal{L}$:
\begin{equation}\label{eq:mse_loss}
    \mathcal{L} = \mathbb{E}_{\mathbb{X}}\Big[ \frac{1}{q \cdot |V| \cdot c} \sum_{t=t_p+1}^{t_p+q} \sum_{i=1}^{|V|} \sum_{j=1}^{c} (\hat{x}_{i, t}[j] - x_{i, t}[j])^2 \Big] 
\end{equation}
Here, $\mathbb{X}$ denotes the distribution of traffic condition data over time, 
$x_{i,t}[j]$ denotes the ground truth value of traffic condition $j$ at vertex $v_i$ and time $t$, while $\hat{x}_{i, t}[j]$ denotes the corresponding model predicted value. 

\subsection{Spatial Correlation Modeling}\label{subsec:spatial}

Motivated by the strong performance of GCNs on general graph prediction problems, we design a GCN variant named LPGCN that is specialized for modeling the spatial correlations  among the traffic conditions of different vertices in a road network graph. 

\textbf{Limitations of standard GCN for traffic prediction.} 
The basic idea of GCN is to compute a  graph-structure-based convolution $\mbox{Conv}^{(h)}(\boldsymbol{X})$ to aggregate the information stored with the graph vertices:
\begin{equation}\label{eq:gcn}
    \mbox{Conv}^{(h)}(\boldsymbol{X}) = \boldsymbol{Z}^{(h)} = \sigma(\boldsymbol{\Tilde{D}}^{-1/2}\boldsymbol{\Tilde{A}}\boldsymbol{\Tilde{D}}^{-1/2}\boldsymbol{Z}^{(h-1)}\boldsymbol{W}^{(h)})
\end{equation}
Here, $h$ denotes a convolution layer number. As $h$ gets larger, the convolution computation output $\boldsymbol{Z}^{(h)}$  preserves higher-order relationships in the graph. Input $\boldsymbol{X}$ is a matrix of vertex features, e.g., $\boldsymbol{X'_{t_i}}$ (i.e., the traffic information at time $t_i$), and $\boldsymbol{Z}^{(0)} = \boldsymbol{X}$. Matrix $\boldsymbol{\Tilde{A}} =  \boldsymbol{A} + \boldsymbol{I_{|V|}}$, where $\boldsymbol{I_{|V|}} \in \mathbb{R}^{|V|\times|V|}$ is the identity matrix, $\boldsymbol{A} \in \mathbb{R}^{|V|\times|V|}$ is the adjacency matrix, and 
$\boldsymbol{\Tilde{D}}$ is a diagonal matrix ($\boldsymbol{\Tilde{D}}_{i, i} = \sum_j\boldsymbol{\Tilde{A}}_{i,j}$). Thus, $\boldsymbol{\Tilde{D}}^{-1/2}\boldsymbol{\Tilde{A}}\boldsymbol{\Tilde{D}}^{-1/2}$ is a normalized adjacency matrix. Matrix $\boldsymbol{W}^{(h)} \in \mathbb{R}^{d \times d}$ contains parameters which are learned by the network. Function $\sigma(\cdot)$ is the activation function (e.g., sigmoid or ReLU).

The adjacency matrix $\boldsymbol{A}$ in Equation~\ref{eq:gcn} enables the (traffic condition) information of a vertex to be passed to its neighbors, while matrix $\boldsymbol{W}$ learns the impact (i.e., weight) of the information from the neighbors. 
Computing Equation~\ref{eq:gcn} for $h$ layers, we can learn the spatial correlations of  the traffic conditions among $h$-hop neighboring vertices. 

Determining the optimal value of $h$ for traffic prediction is difficult. Traffic patterns may be correlated among not only places nearby but also places far away as shown in Fig.~\ref{fig:spatial_distribution_pems04}. 
A small $h$ value only learns the correlation among vertices nearby, while a large $h$ value may produce highly aggregated $\boldsymbol{Z}^{(h)}$ and   
lose the lower-order relationships.  

\textbf{LPGCN.} 
To address the limitations above, we follow a previous model \emph{PGCN}~\cite{zhuang2018dual} and replace matrix $\boldsymbol{\Tilde{A}}$ with another matrix 
to represent the connection between more than just the immediate neighbors.
PGCN replaces $\boldsymbol{\Tilde{A}}$ with a matrix $\boldsymbol{P}$ to represent the vertex connectivity based on shared random walk paths over the vertices.
This results in  the following convolution computation: 
\begin{equation}\label{eq:pgcn}
    \mbox{Conv}_{P}^{(h)}(\boldsymbol{X}) = \boldsymbol{Z}^{(h)} = \sigma(\boldsymbol{D}^{-1/2}\boldsymbol{P}\boldsymbol{D}^{-1/2}\boldsymbol{Z}^{(h-1)}\boldsymbol{W}^{(h)})
\end{equation}

Here, a diagonal matrix $\boldsymbol{D}$ ($\boldsymbol{D}_{i, i} = \sum_j \boldsymbol{P}_{i,j}$) is used for normalization, while the other symbols have the same meanings as those in Equation~\ref{eq:gcn}. Matrix $\boldsymbol{P}$ was calculated as the \emph{point-wise mutual information} (PMI) matrix in the PGCN:  
\begin{equation}
\begin{array}{l}
\displaystyle \boldsymbol{P}_{i,j} = \max\{\log\frac{\beta_{i,j}}{\beta_{i,*} \cdot \beta_{*,j}},0\}\\
\displaystyle \beta_{i,j} = \frac{\boldsymbol{F}_{i,j}}{\sum_{i,j}\boldsymbol{F}_{i,j}}, \quad 
 \beta_{i,*} = \frac{\sum_{j}\boldsymbol{F}_{i,j}}{\sum_{i,j}\boldsymbol{F}_{i,j}},  \quad
\beta_{*,j} = \frac{\sum_{i}\boldsymbol{F}_{i,j}}{\sum_{i,j}\boldsymbol{F}_{i,j}}
\end{array}
\label{eq:ppmi} 
\end{equation}
where $\boldsymbol{F}_{i,j}$ is the frequency of vertex $v_i$ falling on  random walk paths starting from vertex $v_j$. 

Since $\boldsymbol{F}$ is  constrained by the graph structure (i.e., the transition matrix) and the parameters (path length and number of walks per vertex) of random walks, 
$\boldsymbol{P}$ may still fail to capture the correlations for some vertices. 
Vertex $v_i$ far away from vertex $v_j$ may not be reached by any random walks starting from $v_j$, and hence $\boldsymbol{F}_{i,j} = 0$.

To address the limitations of PGCN, we propose to replace $\boldsymbol{F}$ with a learned  matrix $\boldsymbol{\hat{F}}$. 
We initialize $\boldsymbol{\hat{F}}$ with random integer values in $[0, |V|]$. 
This matrix  is later learned  in the model training process to encode the latent connectivity (i.e., spatial correlations) among the vertices. In this way, the spatial correlations learned are not limited to just among vertices connected directly in the graph, but also cover those in a larger distance.  
This leads to smaller traffic prediction errors as shown in our experimental results (Section~\ref{subsec:LPGCN_effect}).  Since our model uses a \emph{learned PMI matrix} $\boldsymbol{\hat{P}}$ (calculated by replacing $\boldsymbol{F}$  with $\boldsymbol{\hat{F}}$ in Equation~\ref{eq:ppmi}), we call it \emph{LPGCN}. 
The  convolution computation of LPGCN is:
\vspace{-1mm} 
\begin{equation}\label{eq:lpgcn}
    \mbox{Conv}_{LP}^{(h)}(\boldsymbol{X}) = \boldsymbol{Z}^{(h)} = \sigma(\boldsymbol{\hat{P}}\boldsymbol{Z}^{(h-1)}\boldsymbol{W}^{(h)})
 \vspace{-1mm}
\end{equation}
Here, we do not need matrix $\boldsymbol{D}$ any more because the normalization is captured by the learned matrix $\boldsymbol{\hat{P}}$ implicitly. 

When adjacency matrices are available, we combine our convolution layer with a \emph{diffusion convolution layer} which has been shown to be effective for traffic prediction~\cite{li2017diffusion}.  
The combined LPGCN convolution layer becomes:
\vspace{-1mm}
\begin{equation}
\begin{array}{l}
\mbox{Conv}_{LP\_A}^{(h)}(\boldsymbol{X}) = \boldsymbol{Z}^{(h)}= 
\sigma\big(\boldsymbol{S}+ \boldsymbol{\hat{P}}\boldsymbol{Z}^{(h-1)}\boldsymbol{W}^{(h)}\big)\vspace{2mm}
\\
\boldsymbol{S} = \sum^K_{i=0}\big(\boldsymbol{Q}^i_{\boldsymbol{f}}\boldsymbol{Z}^{(h-1)}\boldsymbol{W_f}^{(h)} + \boldsymbol{Q}^i_{\boldsymbol{b}}\boldsymbol{Z}^{(h-1)}\boldsymbol{W_b}^{(h)}\big) 
\end{array}
\label{eq:DGCN_LP} 
\end{equation}
Here, $\boldsymbol{Q_f} = \boldsymbol{A}/\sum_j\boldsymbol{A}_{i, j}$ and $\boldsymbol{Q_b}=\boldsymbol{A}^T/\sum_j\boldsymbol{A}_{i, j}^T)$ are the forward and backward transition matrices of a directed graph (only one transition matrix is used for undirected graphs). 
Matrices $\boldsymbol{Q}^i_{\boldsymbol{f}}$  (e.g., $\boldsymbol{Q}^2_{\boldsymbol{f}} = \boldsymbol{Q_f}\cdot \boldsymbol{Q_f}$) and $\boldsymbol{Q}^i_{\boldsymbol{b}}$ contain the $i$-hop transition probabilities between the vertices. Intuitively, $\boldsymbol{S}$ captures correlations among 
neighbors within $K$-hops  in each convolution layer, where $\boldsymbol{W_f}^{(h)}$ and $\boldsymbol{W_b}^{(h)}$ are the parameter matrices to be learned. Using $\boldsymbol{S}$ strengthens  LPGCN in capturing the spatial correlations among close neighbors, while $\boldsymbol{\hat{P}}$ can focus more on distant vertex pairs. 

\subsection{Temporal Correlation Modeling}\label{subsec:temporal}

Next, we learn the temporal correlation of the traffic data from different time points. 

Recall that the mapping output of the linear layers over the input traffic data $\boldsymbol{X_{t_1}}, \boldsymbol{X_{t_2}}, \ldots, \boldsymbol{X_{t_p}}$ is $\boldsymbol{X'_{t_1}}, \boldsymbol{X'_{t_2}}, \ldots, \boldsymbol{X'_{t_p}}$. 
To learn the temporal correlations among these matrices, ideally we shall examine the joint impact of all possible combinations of these matrices on a matrix $\boldsymbol{\hat{X}_{t_{p+i}}}$ to be predicted. 
Given $p$ input matrices, there are a total of $2^{p}$ combinations, which will be  too expensive to compute as $p$ becomes large. 
For computation efficiency, we focus on combinations of matrices adjacent in time. 

\textbf{Multi-path temporal convolution.} 
We feed  the matrices $\boldsymbol{X'_{t_1}}, \boldsymbol{X'_{t_2}}, \ldots, \boldsymbol{X'_{t_p}}$ into $p-1$ 1-dimension CNNs (i.e., a multi-path CNN) with increasing kernel sizes (i.e., $2, 3, \ldots, p$) and stride size 1 separately. 
Using a 1-dimension CNN with kernel size $j$ (denoted by $\mbox{CNN}_j$), we aggregate every $j$ matrices adjacent in the list $\boldsymbol{X'_{t_1}}, \boldsymbol{X'_{t_2}}, \ldots, \boldsymbol{X'_{t_p}}$ to learn their combined temporal impact, which yields a sequence of $p-j+1$ matrices, denoted by 
$\boldsymbol{\overline{X}_{j, 1}}, \boldsymbol{\overline{X}_{j, 2}}, \ldots, \boldsymbol{\overline{X}_{j, p-j+1}}$. For $j \in [2, p]$: 
\begin{equation}
\begin{array}{l}
     \boldsymbol{\overline{X}_{2, 1}},\boldsymbol{\overline{X}_{2, 2}},\ldots, 
     \boldsymbol{\overline{X}_{2, p-1}} = \mbox{CNN}_2(\boldsymbol{X'_{t_1}}, \boldsymbol{X'_{t_2}}, \ldots, \boldsymbol{X'_{t_p}})
     \\
     \boldsymbol{\overline{X}_{3, 1}},\boldsymbol{\overline{X}_{3, 2}},\ldots, \boldsymbol{\overline{X}_{3, p-2}}= \mbox{CNN}_3(\boldsymbol{X'_{t_1}}, \boldsymbol{X'_{t_2}}, \ldots, \boldsymbol{X'_{t_p}})\\
\ldots\\
     \boldsymbol{\overline{X}_{p, 1}}= \mbox{CNN}_{p}(\boldsymbol{X'_{t_1}}, \boldsymbol{X'_{t_2}}, \ldots, \boldsymbol{X'_{t_p}})
\end{array}
\label{eq:cnn}
\end{equation} 
Here, every output matrix $\boldsymbol{\overline{X}_{j, i}}$ is in $\mathbb{R}^{|V|\times d}$. 

We use all these output matrices to predict the traffic conditions at $t_{p+1}, t_{p+2}, \ldots, t_{p+q}$ in parallel instead from one time point to another. This way, we reduce the computation time while avoiding the error propagation issue caused by using prediction results of the immediate future for predicting further into the future.

We concatenate these matrices (and  $\boldsymbol{X'_{t_1}}, \boldsymbol{X'_{t_2}}, \ldots, \boldsymbol{X'_{t_p}}$). Let the  concatenated matrix be $\boldsymbol{\overline{X}} \in \mathbb{R}^{(1+p)p/2 \times |V|\times d}$. 
\begin{equation}
\begin{split}
\boldsymbol{\overline{X}} = &\boldsymbol{X'_{t_1}} \oplus \boldsymbol{X'_{t_2}} \oplus \ldots \oplus \boldsymbol{X'_{t_p}} \oplus \\
&\boldsymbol{\overline{X}_{2, 1}}\oplus \boldsymbol{\overline{X}_{2, 2}} \oplus \ldots\oplus
   \boldsymbol{\overline{X}_{2, p-1}}\oplus \\
   &\ldots\\
   &\boldsymbol{\overline{X}_{p, 1}}
   \end{split} 
\end{equation}
Here, `$\oplus$' denotes the concatenation operation.

\textbf{Temporal attention.} When making prediction for time $t_{p+1}$, we use $t_{p+1}$  to compute an attention over $\boldsymbol{\overline{X}}$: 
\begin{equation}
\begin{split}
   \boldsymbol{T_{t_{p+i}}} = softmax(\boldsymbol{E_{t_{p+i}}}) \cdot \boldsymbol{\overline{X}}
\end{split}
\label{eq:attention} 
\end{equation}
Here, $softmax(\cdot)$ is the softmax function, $\boldsymbol{E_{t_{p+i}}} \in \mathbb{R}^{|V|\times p(p+1)/2}$ are the embeddings of $t_{p+1}$ to be multiplied with the vectors of $|V|$ different vertices in $\boldsymbol{\overline{X}}$, and the multiplication is denoted by `$\odot$' in Fig.~\ref{fig:model}c. Here, we use a different embedding of $t_{p+1}$ for each different vertex such that each vertex has a different time-based attention. 

We generate $\boldsymbol{E_{t_{p+i}}}$ as follows. To encode the periodic traffic patterns, we first represent $t_{p+1}$ with two one-hot vectors, one encoding the time of day and the other encoding the day of week of $t_{p+1}$.  The time-of-day one-hot vector has $12\times 24 = 288$ dimensions where each dimension represents  a five-minute interval. The day-of-week one-hot vector  has 7 dimensions where each dimension represents a different day in a week (a public holiday can be treated either as Sunday or an additional day type). The two vectors are concatenated together and fed into linear layers (one layer of $|V| \cdot p(p+1)/2$ nodes in our experiments), which map them into a latent time space and generates $\boldsymbol{E_{t_{p+i}}}$. In this latent space, time points with similar traffic patterns are expected to have embeddings that are close to each other. 

\textbf{Discussion.} Our use of a multi-path CNN differs from existing traffic prediction models that  use multiple CNNs for temporal correlation learning such as \emph{Graph WaveNet}~\cite{wu2019graph} and \emph{STGCN}~\cite{yu2017spatio}. As mentioned in Section~\ref{sec:lit_review},  these models use two CNNs with different activation functions, sigmoid and tanh. The output of both CNNs is multiplied, such that the sigmoid CNN re-scales its output to range $(0,1)$ and works as a gate to control the strength of  the signals from the tanh CNN that go  through the gate, while each of our CNN paths computes the impact of traffic observations from a different past time combination, and all output from different CNN paths is weighted by an attention mechanism computed based on the prediction time.

The \emph{AttConv} model~\cite{yin-schutze-2018-attentive} has combined the attention mechanism with CNNs for natural language processing. It differs from our idea in that it takes the output of an attention computation (on a separate input called the ``context'') to modify the input of a CNN, while we compute attention on the output of multiple CNN paths to aggregate them.  
The \emph{SKNets} model~\cite{DBLP:conf/cvpr/LiW0019} uses the attention mechanism with a multi-path CNN for image classification. It computes attention weights from intermediate output of a multi-path CNN, which are multiplied with the output of the CNN paths to generate the final multi-path CNN output. We compute attention weights from outside the multi-path CNN instead. 
The  \emph{attentive group equivariant convolutional network}~\cite{pmlr-v119-romero20a} combines the attention mechanism with CNNs, also for image classification. It uses two CNN groups, each consisting of four CNNs with rotating filters. One group yields the attention (using the softmax function) while the other yields the signals, much like the gated mechanism to combine CNNs as done in Graph WaveNet and STGCN discussed above, which is different from our idea.

\subsection{Gated Fusion}\label{subsec:gatefusion}

We follow Zheng et al.~\cite{zheng2020gman} and use a gated fusion mechanism to fuse the output of the LPGCNs with the temporal attention as follows. 
\vspace{-1mm}
\begin{equation}
\begin{array}{l}
    \quad \quad \quad \quad \boldsymbol{\hat{X}'_{t_{p+i}}} = \alpha \cdot \boldsymbol{Z_{t_i}} + (1 - \alpha) \cdot \boldsymbol{T_{t_{p+i}}} \\ 
    \quad \quad \quad \quad \alpha = \sigma(\boldsymbol{Z_{t_i}} \cdot \boldsymbol{W_1} + \boldsymbol{T_{t_{p+i}}} \cdot  \boldsymbol{W_2} + \boldsymbol{b})
\end{array}
\label{eq:gate_fusion}
\vspace{-1mm}
\end{equation}
Here, $\boldsymbol{\hat{X}'_{t_{p+i}}} \in \mathbb{R}^{|V| \times d}$ denotes the fusion output, $\boldsymbol{Z_{t_i}}$ denotes the LPGCN output given the traffic condition at $t_i$ (i.e., $\boldsymbol{X'_{t_i}}$) as the input, $\boldsymbol{T_{t_{p+i}}}$ denotes the temporal attention output for time $t_{p+i}$, $\sigma$ denotes the sigmoid activation function, and $\boldsymbol{W_1}$, $\boldsymbol{W_2}$, and $\boldsymbol{b}$ are learnable parameters.
Note that we have used the spatial correlations at time $t_i$ (i.e., $\boldsymbol{Z_{t_i}}$) to make predictions  for time $t_{p+i}$, because this shows better empirical performance than  using only the spatial correlations at time $t_p$ (i.e., the latest observation time) to make predictions for all future time points.  We conjecture that this approach offers extra input variability to help generate different prediction output for different time points. When $\boldsymbol{Z_{t_i}}$ is unavailable (e.g., due to limited historical data), we can fall back to just using $\boldsymbol{Z_{t_p}}$.

Finally, we feed  $\boldsymbol{\hat{X}'_{t_{p+i}}}$ to linear layers (we use two layers of 10 and $c$ nodes in our experiments) to map the predicted values from $\mathbb{R}^{|V| \times d}$ back to $\mathbb{R}^{|V| \times c}$ to generate the traffic prediction $\boldsymbol{\hat{X}_{t_{p+i}}}$ for time $t_{p+i}$.

\section{Experiments}\label{sec:exp}

We present experimental settings in Section~\ref{subsec:exp_settings} and overall performance  comparison  results in Section~\ref{subsec:exp_results}.  We study the effectiveness of the spatial and temporal correlation modules in Sections~\ref{subsec:LPGCN_effect} and~\ref{subsec:CNN_effect}, respectively. 
We test the model extendibility  by integrating it with GAN in Section~\ref{subsec:exp_gan}.

\subsection{Settings}\label{subsec:exp_settings}
We run experiments on a High Performance Computing (HPC) system named \emph{Spartan} ({\url{https://dashboard.hpc.unimelb.edu.au}) that offers a virtual NVIDIA Tesla V100 GPU (15 GB GPU memory) to each user. The models are implemented in Python 3.6 and Tensorflow 2.7.

\textbf{Datasets.}
We use two real datasets from PeMS in two different districts. \textbf{PEMS04} contains traffic speed data collected by sensors on highways of District 4, and \textbf{PEMS08} contains that of District 8. Both datasets are collected during a 2-month period between 1st of December 2020 and 31st of January 2021. 
The speed data is aggregated and reported by the PeMS system at 5-minute intervals. 
Thus, each  dataset contains speed data at 17,856 intervals (i.e., 12 intervals per hour $\times$ 24 hours per day $\times$ 62 days). For each district, we randomly sampled 400 sensors. The road networks are collected from OpenStreetMap (\url{https://planet.osm.org}).
We divide the datasets (by time in ascending order) into training (70\%), validation (10\%), and testing (20\%) subsets. 
Following previous studies~\cite{zheng2020gman,li2017diffusion,yu2017spatio,wu2019graph}, we take the traffic condition (speed) data  at 5-minute intervals for the past hour as the input and predict the traffic condition for the coming hour (i.e., $p = q = 12$). 
Since we only collected the traffic speed data in the datasets, the number of traffic conditions of interest, $c$, is $1$ in our experiments. \\

\textbf{Competitors.}
We compare our proposed model \textbf{\model} with the following models:

(1) \textbf{ARIMA}~\cite{williams2003modeling} (Auto-Regressive Integrated Moving Average) is a commonly used non-neural-network based traffic prediction model. 
    
(2) \textbf{FC-LSTM}~\cite{li2017diffusion} combines a fully connected feedforward network  with an LSTM network for traffic prediction. 

(3) \textbf{WaveNet}~\cite{oord2016wavenet} is a gated convolutional network for sequence data prediction. 
    
(4) \textbf{DCRNN}~\cite{li2017diffusion} 
combines DGCN and RNN in an auto-encoder architecture for traffic prediction. 
    
(5) \textbf{STGCN}~\cite{yu2017spatio} 
stacks graph convolution layers  with gated 1-D convolutional  networks for spatio-temporal correlation modeling and traffic prediction.
                 
(6) \textbf{Graph WaveNet}~\cite{wu2019graph} 
combines the WaveNet~\cite{oord2016wavenet} and a GCN for traffic prediction. 
     
(7) \textbf{GMAN}~\cite{zheng2020gman} is the state-of-the-art model for traffic (speed)  prediction. It uses attention networks to model the spatial and temporal correlations and an encoder-decoder framework for traffic prediction.

(8) \textbf{ASTGCN}~\cite{guo2019attention} uses attention with GCN to model the spatial correlation and uses attention with CNN to model the temporal correlation, for traffic (flow) prediction.  
 
(9) \textbf{AGCRN}~\cite{DBLP:conf/nips/0001YL0020} proposes an adaptive GCN,  which resembles the adaptive adjacency matrix idea of Graph WaveNet, and combines it with GRU for traffic prediction. 
     
(10) \textbf{STSGCN}~\cite{DBLP:conf/aaai/SongLGW20} proposes a  \emph{spatial-temporal synchronous graph convolutional} (STGC) module to learn localized spatial and temporal correlations among vertices in three consecutive time points. It builds a hierarchy of such modules for spatio-temporal network data prediction. 
        
(11) \textbf{STFGNN}~\cite{DBLP:conf/aaai/LiZ21} adds temporal graphs and temporal connectivity graphs into the STGC module and combines it with gated CNNs for traffic (flow) prediction.

\textbf{Implementation details.} 
For DCRNN, STGCN, Graph WaveNet, GMAN, ASTGCN, STSGCN, and STFGNN, 
we used the original source code and default settings from the respective  papers. 
The AGCRN original source code has a compatibility issue with CUDA and PyTorch libraries. We use an implementation from LibCity (\url{https://github.com/LibCity/Bigscity-LibCity}) instead. 
The other models do not have source code directly available. For WaveNet, we obtain an implementation via adapting the Graph WaveNet source code to disable the GCN component. We use the default settings of Graph WaveNet for WaveNet. For ARIMA, we use the implementation from a Python 3  package named \emph{statsmodels} and follow the settings used in the  DCRNN paper~\cite{li2017diffusion}. We implement FC-LSTM following the description and settings in the  DCRNN paper.  

For our model \model, 
we set  $d = 100$ (i.e., the dimensionality of the mapped space of the input) and 
 $K = 2$ in LPGCN  (we use one LPGCN convolution layer). 
 We use the \emph{AdamOptimizer} with a learning rate of 0.0001, batch size of 4, and 20 epochs with early stopping in all experiments. We used grid search to learn these optimized hyper-parameters from $d \in \{20, 60, 100, 140, 180\}$,  $K \in \{2, 4, 6, 8, 10\}$, learning rate among $\{0.001, 0.0003,  0.0001\}$, and batch size among $\{64, 32, 16, 8, 4\}$. Source code of our model is at: \url{https://github.com/alvinzhaowei/GAMCN}.

\textbf{Evaluation metrics.} 
We use the traffic speed data in the past 60 minutes as the  input to 
predict the traffic speed for up to 60 minutes into the future. 
Following previous studies~\cite{li2017diffusion,zheng2020gman}, we apply the \emph{z-score normalization} to our input data using equation 
    $A_{norm}= \frac{A - \mu}{\sigma}$.  
Here, $A$ is an original traffic speed value, $\mu$ is traffic speed mean, $\sigma$ is the standard deviation, and $A_{norm}$ is the normalized value. After this data normalization, there is no need to apply an activation function in our output layer.

We measure prediction errors by the \emph{Mean Absolute Error} (\textbf{MAE}), \emph{Root Mean Square Error} (\textbf{RMSE}), and \emph{Mean Absolute Percentage Error} (\textbf{MAPE}), at 15, 30, and 60 minutes.

\subsection{Overall Model Prediction Performance}\label{subsec:exp_results}

\begin{table*}[ht]
\centering
    \caption{Prediction Errors on the PEMS04 and PEMS08 Datasets}
    \label{tab:error}
               \vspace{-2mm} 
    {\begin{tabular}{l|l|rrr|rrr|rrr}
        \toprule
        \multirow{2}{*}{Dataset} & \multirow{2}{*}{Model} & \multicolumn{3}{c|}{15 minutes} & \multicolumn{3}{c|}{30 minutes}& \multicolumn{3}{c}{60 minutes}\\ \cline{3-11}
         & & MAE  & MAPE & RMSE & MAE  & MAPE & RMSE  & MAE  & MAPE & RMSE\\
        \midrule
        \midrule
\multirow{8}{*}{PEMS04}

& ARIMA& 1.29 & 1.97\%& 2.69 & 1.73 & 3.20\% & 5.41 & 2.15 & 3.70\% & 4.22\\ \cline{2-11} 
& FC-LSTM& 1.37 & 2.03\%& 2.81 & 1.41 &  2.69\% &  3.80 & 1.44 & 2.59\% & 3.53\\ \cline{2-11} 
& WaveNet& 1.21 & 1.88\%& 2.39 & 1.39 & 2.65\% & 2.82 & 1.56 & 3.15\% & 3.82\\ \cline{2-11} 
& DCRNN& 1.15 & 1.80\%& 2.28 & 1.25 & 2.40\% & 2.55 & 1.41 & 2.53\% & 3.17\\ \cline{2-11} 
& STGCN& 1.19 & 1.82\%& 2.34 & 1.37 & 2.57\% & 2.74 & 1.48 & 2.61\% & 3.32\\ \cline{2-11} 

& ASTGCN & 1.17 & 1.80\% & 2.33 & 1.37 & 2.59\% & 2.77 & 1.50 & 2.67\% & 3.41 \\ \cline{2-11}
& AGCRN & 1.14 & 1.77\% & 2.21 & 1.22 & 2.33\% & 2.51 & 1.38 & 2.48\% & 3.14 \\ \cline{2-11}
& STSGCN & 1.08 & 1.82\% & 2.20 & 1.24 & 2.32\% & 2.66 & 1.31 & 2.32\% & 3.00 \\ \cline{2-11}
& STFGNN & 1.03 & 1.79\% & 2.11 & 1.21 & 2.27\% & 2.58 & 1.27 & 2.25\% & 2.95 \\\cline{2-11}

& Graph WaveNet & 0.99 & 1.72\% & 1.85 & 1.16 &2.15\% & 2.43 & 1.32 &2.31\% & 3.02 \\ \cline{2-11}
& GMAN & 1.03 & 1.73\% & 1.91 & 1.15 & 2.13\% & 2.43 & 1.23 & 2.21\% & 2.97 \\  \cline{2-11}

 & \textbf{\model\ (proposed)} & \textbf{0.96} & \textbf{1.71\%}  & \textbf{1.80} & \textbf{1.12} & \textbf{2.05\%} & \textbf{1.97}  & \textbf{1.21} & \textbf{2.14\%} & \textbf{2.68}\\ \bottomrule
 
 \multirow{8}{*}{PEMS08}
 & ARIMA& 1.62 & 4.02\%& 3.70 & 2.19 & 6.33\% & 4.43 & 2.95 & 6.10\% & 4.43\\ \cline{2-11} 
& FC-LSTM& 1.55 & 3.80\%& 3.49 & 1.77 &  4.37\% &  3.72 & 1.89 &  4.77\% &  5.32\\ \cline{2-11} 
& WaveNet& 1.27 & 2.53\%& 2.91 & 1.57 & 4.05\% & 3.34 & 1.84 & 4.88\% & 5.27\\ \cline{2-11} 
& DCRNN& 1.25 & 2.49\%  & 2.90 & 1.53 & 3.88\% & 3.14 & 1.81 & 4.68\% & 5.14\\ \cline{2-11} 
& STGCN& 1.28 & 2.53\%& 2.95 & 1.57 & 3.97\% & 3.32 & 1.91 & 4.95\% & 5.41\\ \cline{2-11} 

& ASTGCN & 1.29 & 2.57\% & 3.03 & 1.59 & 4.01\% & 3.36 & 1.92 & 5.01\% & 5.52  \\ \cline{2-11}
& AGCRN & 1.25 & 2.51\% & 2.92 & 1.55 & 3.86\% & 3.16 & 1.79 & 4.62\% & 5.07  \\ \cline{2-11}
& STSGCN & 1.26 & 2.50\% & 2.92 & 1.52 & 3.72\% & 3.07 & 1.68 & 4.31\% & 3.93  \\ \cline{2-11}
& STFGNN & 1.25 & 2.51\% & 2.84 & 1.49 & 3.66\% & 3.04 & 1.65 & 4.27\% & 3.85  \\ \cline{2-11}

& Graph WaveNet & 1.23 & 2.42\% & 2.74 & 1.48 & 3.67\% & 2.99 & 1.70 & 4.28\% & 4.01  \\ \cline{2-11}
& GMAN & 1.24 & 2.43\% & 2.87 & 1.45 & 3.48\% & 2.92 & \textbf{1.61} & 4.10\% & 3.53\\  \cline{2-11}
 & \textbf{\model\ (proposed)} & \textbf{1.22} & \textbf{2.41\%}  & \textbf{2.65} & \textbf{1.43} & \textbf{3.43\%} & \textbf{2.87}& \textbf{1.61} &  \textbf{4.03\%} & \textbf{3.31} \\ 
        \bottomrule

        \end{tabular}}
        \vspace{-4mm}
\end{table*}

We first report overall  model  prediction error and efficiency.

\textbf{Prediction errors.}
As Table~\ref{tab:error} shows, on both datasets, our model \model\ yields the smallest errors when predicting the traffic condition (i.e., traffic speed) for  15, 30, and 60 minutes ahead.  
\model\ outperforms the competitors by up to  45.4\% in MAE (\model\ vs. ARIMA on PEMS08 for 60 minutes), 45.8\% in MAPE  (\model\ vs. ARIMA on PEMS08 for 30 minutes), and  63.6\% in RMSE (\model\ vs. ARIMA on PEMS04 for 30 minutes). Comparing with the state-of-the-art model GMAN, 
\model\ reduces the MAE, MAPE, and RMSE by up to 6.8\% (on PEMS04 for 15 minutes), 3.8\% (on PEMS04 for 30 minutes), and 18.9\% (on PEMS04 for 30 minutes), respectively. This  confirms the effectiveness of \model\ to produce highly accurate predictions, via capturing the spatial and temporal correlations among the traffic speed observations at different places and time.  

GMAN has closer results (same MAE on PEMS08) to those of \model\ when predicting for 60 minutes ahead.   
GMAN was designed to predict for a longer time ahead (e.g., an hour). 
It uses a temporal attention mechanism which shows a similar effect (but with different design) to the CNN and attention modules in \model. When  predicting for further ahead, the temporal impact is more significant. Thus, the two models yield a closer performance.
Even in this case, \model\ still outperforms GMAN in MAPE and RMSE by 1.7\% and 6.2\% (on PEMS08), respectively.

Graph WaveNet, on the other hand, is the best baseline model when predicting for a shorter term, i.e., 15 minutes ahead. However, its performance degrades as the predictions are made for further into the future. For example, the improvements of \model\ over Graph WaveNet on PEMS04 in MAE, MAPE, and RMSE grows from 3.0\%, 0.5\%, and 2.7\% to 8.3\%, 7.4\%, and 11.3\%, respectively, when the prediction time increases from 15 minutes to 60 minutes. 
These observations confirm the robustness of \model\  in making highly accurate predictions for both the immediate and a longer time into the future. 

ASTGCN, STSGCN, and STFGNN use GCN  and adjacency matrices directly while \model\ uses LPGCN to  learn an adjacency matrix, which shows a stronger spatial correlation modeling capability.  
AGCRN also learns an adaptive  adjacency matrix but uses a simpler approach which is similar to that in Graph WaveNet and is shown to be less effective. Meanwhile, the multi-path CNN and temporal attention modules of \model\ allow it to effectively  learn the joint impact of the past traffic observations and the prediction time (i.e., the temporal correlations). These explain for the overall lower prediction errors of \model. 

To be fair, we note that ASTGCN, AGCRN, STSGCN, and STFGNN were designed for and/or tested only on traffic flow data originally. 
They may not perform the best on traffic speed data (which is used in our experiments). Still, STSGCN and STFGNN are among the strongest when predicting for 
60 minutes ahead, where they are only worse than GMAN and \model\  on both datasets, due to their spatio-temporal graph structures that learn the spatial and temporal correlation patterns together. Designing a model that achieves state-of-the-art results for both traffic flow and speed data would be an interesting future work. 

\begin{table}[ht]
\vspace{-2mm}
\centering
\caption{Running Times on the PEMS04 Dataset (Second)}
\label{tab:tc}
\vspace{-2mm}
\setlength{\tabcolsep}{0.6mm}
\begin{tabular}{l|r|r|r}
\toprule
Model     & Training time & Prediction time  & Prediction time  \\
     &  (per epoch) &  (per time point) & (12 time points) \\
\midrule
\midrule
ARIMA & 119.6  & \textbf{0.6}& \textbf{7.2}   \\\hline
FC-LSTM & \textbf{62.5}& 21.7 & 259.0\\\hline
WaveNet & 462.5.& 36.2& 36.2\\\hline
DCRNN & 1213.5  & 51.1& 615.9   \\\hline
STGCN & 101.4  & 43.7 & 523.2      \\\hline

ASTGCN  &624.1 &	59.7	& 59.7  \\\hline
AGCRN & 638.3 &	221.9 & 221.9  \\\hline
STSGCN  & 422.4 &	34.8 &	34.8  \\\hline
STFGNN  & 548.5 &	52.6	 & 52.6  \\\hline

Graph WaveNet & 485.5   & 45.9  & 45.9 \\\hline
GMAN & 519.7   & 50.5  & 50.5  \\\hline

\textbf{\model\ (proposed)} & 479.9  & 38.7 & 38.7  \\
\bottomrule
\end{tabular}
\vspace{-2mm}
\end{table}

\textbf{Model time efficiency.}  
Table~\ref{tab:tc} further shows the training and prediction times of the models. For succinctness, we only report the results on the PEMS04 dataset. Since both datasets have the same size, the comparative model performance on PEMS08 has a similar pattern. 

We measure the per-epoch training time following the Graph WaveNet and  GMAN papers (the models use early termination and have randomness in the full model training time) for all models except ARIMA. We report the  full model training time of ARIMA 
since its \emph{statsmodels} implementation does not support per-epoch running time measurement.  Also note that the training of ARIMA is run on the CPU because  this  implementation  does not support GPUs. 

Among the deep learning models, FC-LSTM and STGCN have the fastest training time, due to their relatively simple sequence-to-sequence structures. However, as shown above, their prediction errors are worse than many other models. 
The other models, which do not use RNN in their structures, all have training times in a similar scale, including WaveNet, Graph WaveNet, ASTGCN, STSGCN, STFGNN, GMAN, and \model. 
 Our model \model, in particular, reduces the training time by 7.7\% comparing with the state-of-the-art model GMAN while outperforming it in the prediction errors. 
DCRNN has the longest training time because it needs to train multiple RNN modules. AGCRN uses GRU which also lacks efficiency. 

In terms of prediction time, ARIMA is the fastest due to its simple procedure. All other models make predictions for a time point (e.g., 15 minutes) using between 21 and 60 seconds (AGCRN uses a GRU and takes 221.9 seconds to predict for 12 time points together). 
Our model \model\ makes predictions 23.4\% faster than GMAN does. It is also faster than most of the other deep learning models except for  
FC-LSTM and WaveNet which have simpler structures, and STSGCN which has a  similar  complexity in the structure. 

Table~\ref{tab:tc} also shows the times taken by the models to make predictions for 12 time points (i.e., for 60 minutes at every 5-minute interval). The prediction times do not increase for WaveNet, ASTGCN, AGCRN, STSGCN, STFGNN, Graph~WaveNet, GMAN, and our \model\ model, since these models make  predictions for the 12 time points together. The other models, i.e., ARIMA, FC-LSTM, DCRNN, and STGCN, predict progressively from one time point to another. Their prediction times increase with more time points considered. These results confirm the time efficiency of our \model\ model which makes it practical for real-world applications with high efficiency requirements. 

\subsection{Effectiveness of the Spatial Correlation Module}\label{subsec:LPGCN_effect}

Next, we study the effectiveness our LPGCN module (Section~\ref{subsec:spatial}). We compare  \model\  with three variants, namely, \textbf{\model-GCN}, \textbf{\model-DGCN}, and \textbf{\model-PGCN}, where the LPGCN module is replaced by the vanilla GCN, DGCN~\cite{li2017diffusion} (i.e., removing $\boldsymbol{\hat{P}}\boldsymbol{Z}^{(h-1)}\boldsymbol{W}^{(h)}$ from Equation~\ref{eq:DGCN_LP}), and  PGCN~\cite{zhuang2018dual} (PGCN uses random walks with path length $|V|$, 10 walks per vertex, and edge weights to form the transition probabilities), respectively.

\begin{table}[ht]
    \vspace{-2mm}
\setlength{\tabcolsep}{0.8mm}
\centering
    \caption{Prediction Errors with and without LPGCN (RMSE results omitted for space limit)}
    \label{tab:gcns}
              \vspace{-2mm}
    {\begin{tabular}{l|l|rr|rr|rr}
        \toprule
        \multirow{2}{*}{Dataset} & \multirow{2}{*}{Model} & \multicolumn{2}{c|}{15 minutes} & \multicolumn{2}{c|}{30 minutes}& \multicolumn{2}{c}{60 minutes}\\ \cline{3-8}
         & & MAE  & MAPE &  MAE  & MAPE &  MAE  & MAPE  \\
        \midrule
        \midrule
\multirow{3}{*}{PEMS04}

& \model-GCN & 1.05 & 1.79\% & 1.23 & 2.15\% & 1.30 & 2.22\% \\ \cline{2-8} 
& \model-DGCN & 0.99 & 1.75\%  & 1.17 & 2.11\% & 1.24 & 2.15\%   \\ \cline{2-8}
& \model-PGCN &0.99 & 1.78\%  & 1.14 & 2.08\%  & 1.24 & 2.18\%  \\ \cline{2-8}
 & \textbf{\model} & \textbf{0.96} & \textbf{1.71\%}  & \textbf{1.12} & \textbf{2.05\%} &  \textbf{1.21} & \textbf{2.14\%} \\ \bottomrule
 
 \multirow{3}{*}{PEMS08}

& \model-GCN & 1.35 & 2.51\% & 1.55 & 3.55\% & 1.71 & 4.19\% \\ \cline{2-8}
& \model-DGCN & 1.29 & 2.47\% & 1.47 & 3.50\% & 1.66 & 4.11\%  \\  \cline{2-8}
& \model-PGCN &1.23 & 2.45\% & 1.44 & 3.45\% & 1.62 & 4.08\% \\ \cline{2-8}
 & \textbf{\model} & \textbf{1.22} & \textbf{2.41\%}   & \textbf{1.43} & \textbf{3.43\%} & \textbf{1.61} & \textbf{4.03\%}  \\ 
        \bottomrule

        \end{tabular}}
        \vspace{-2mm}
\end{table}

\textbf{Impact of LPGCN.} As Table~\ref{tab:gcns} shows, comparing with  \model-GCN,  \model\ (using LPGCN) reduces the prediction errors by up to  9.6\% (on PEMS08 for 15 minutes) and 4.7\% (on PEMS04 for 30 minutes) in MAE and MAPE, respectively (similar results are observed on RMSE, which are omitted hereafter due to space limit). Comparing with \model-DGCN, 
\model\ reduces the two error measurements by up to 5.4\% (on PEMS08 for 15 minutes) and  2.8\% (on PEMS04 for 30 minutes), respectively. Comparing with \model-PGCN, 
\model\ reduces the two error measurements by up to 3.0\% and 3.9\% (all on PEMS04 for 15 minutes), respectively. These performance gains confirm the effectiveness of LPGCN in capturing richer spatial correlations, which help reduce the prediction errors.

\begin{figure}[ht]
\vspace{-5mm}
\begin{subfigure}{.520\linewidth}
  \includegraphics[width=1\linewidth]{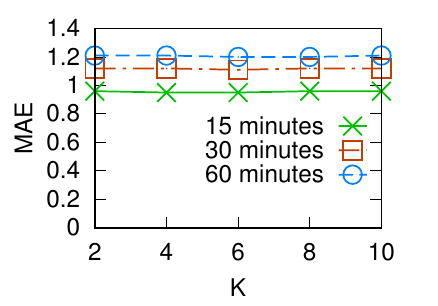}  
  \caption{PEMS04}
  \label{fig:k_pems04_mae}
\end{subfigure}
\hspace{-5mm}
\begin{subfigure}{.52\linewidth}
  \includegraphics[width=1\linewidth]{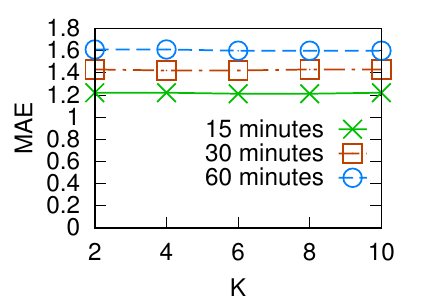}  
  \caption{PEMS08}
  \label{fig:k_pems08_mae}
\end{subfigure}
\vspace{-2mm}
\caption{Impact of $K$ on prediction errors}\label{fig:impact_k}
\vspace{-2mm}
\end{figure}

\textbf{Impact of $K$.}  Fig.~\ref{fig:impact_k} shows the impact of the value of parameter $K$ 
 in Equation~\ref{eq:DGCN_LP} for LPGCN. We find that the value of $K$ has a limited impact on the model performance (in terms of MAE; MAPE and RMSE results are similar and are omitted).  This is because the PMI learning process in LPGCN has already captured both close and distant correlations among the vertices in the graph.  Thus, for computation efficiency, we have used $K=2$ in our other experiments. 

\begin{table}[ht]
    \vspace{-2mm}
\setlength{\tabcolsep}{0.7mm}
\centering
    \caption{Prediction Errors without Given Graph Adjacency Matrices (There is no result DCRNN, STGCN, GMAN, ASTGCN, STSGCN, and STFGNN, because these models require adjacency matrices to learn the spatial correlations)}
    \label{tab:error_na}
               \vspace{-2mm}
    {\begin{tabular}{l|l|rr|rr|rr}
        \toprule
        \multirow{2}{*}{Dataset} & \multirow{2}{*}{Model} & \multicolumn{2}{c|}{15 minutes} & \multicolumn{2}{c|}{30 minutes}& \multicolumn{2}{c}{60 minutes}\\ \cline{3-8}
         & & MAE  & MAPE & MAE  & MAPE & MAE  & MAPE \\
        \midrule
        \midrule
\multirow{8}{*}{PEMS04}

& ARIMA& 1.29 & 1.97\%& 1.73 & 3.20\% & 2.15 & 3.70\% \\ \cline{2-8} 
& FC-LSTM& 1.37 & 2.03\%& 1.41 &  2.69\% & 1.44 & 2.59\% \\ \cline{2-8} 
& WaveNet& 1.21 & 1.88\%&  1.39 & 2.65\%  & 1.56 & 3.15\% \\ \cline{2-8} 

& AGCRN & 1.14 & 1.77\%  & 1.22 & 2.33\%  & 1.38 & 2.48\%  \\ \cline{2-8}

& Graph WaveNet & 1.01 & \textbf{1.73\%}  & 1.17 &2.18\%  & 1.33 &2.32\% \\ \cline{2-8}

 & \textbf{\model} & \textbf{1.00} & \textbf{1.73\%}   & \textbf{1.15} & \textbf{2.06\%} & \textbf{1.22} & \textbf{2.14\%} \\ \bottomrule
 
 \multirow{8}{*}{PEMS08}
 & ARIMA& 1.62 & 4.02\% & 2.19 & 6.33\%  & 2.95 & 6.10\% \\ \cline{2-8} 
& FC-LSTM& 1.55 & 3.80\% & 1.77 &  4.37\% & 1.89 &  4.77\% \\ \cline{2-8} 
& WaveNet& 1.27 & 2.53\% & 1.57 & 4.05\%  & 1.84 & 4.88\% \\ \cline{2-8} 
& AGCRN & 1.25 & 2.51\%& 1.55 & 3.86\%& 1.79 & 4.62\%  \\ \cline{2-8}

& Graph WaveNet & 1.26 & 2.44\%  & 1.51 & 3.68\% & 1.72 & 4.29\%   \\ \cline{2-8}

 & \textbf{\model} & \textbf{1.27} & \textbf{2.43\%}   & \textbf{1.46} & \textbf{3.44\%} & \textbf{1.63} &  \textbf{4.04\%}  \\ 
        \bottomrule
        
 \multirow{8}{*}{MEL}
 & ARIMA& 3.28 & 15.60\%& 3.31 & 15.73\%  & 3.37 & 15.85\% \\ \cline{2-8} 
& FC-LSTM& 2.61 & 12.71\% & 2.85 &  13.79\%  & 2.93 &  13.92\%\\ \cline{2-8} 
& WaveNet& 2.59 & 12.21\%  & 2.83 & 13.53\%  & 2.95 & 14.95\% \\ \cline{2-8} 

& AGCRN & 2.54 & 11.88\% & 2.72 & 12.51\%  & 2.77 & 12.88\% \\ \cline{2-8}

& Graph WaveNet & 2.56 & 11.89\%  & 2.79 & 12.88\%  & 2.92 & 13.48\%  \\ \cline{2-8}

 & \textbf{\model} & \textbf{2.52} & \textbf{11.85\%}   & \textbf{2.68} & \textbf{12.40\%} & \textbf{2.75} &  \textbf{12.84\%}  \\ 
        \bottomrule
        \end{tabular}}
        \vspace{-2mm}
\end{table} 

\textbf{Impact of missing graph adjacency matrices.}
Since LPGCN learns a PMI matrix, it does not rely on a given graph adjacency matrix. This enables \model\ to be used without a given 
adjacency matrix, e.g., when the underlying road network data is unavailable.  Below, we test our model performance under such settings.

We remove the adjacency matrices from PEMS04 and 
PEMS08. We further add a local dataset \textbf{MEL} which contains traffic speed data collected by eight sensors from Melbourne CBD between 1st of January and 30th of June, 2019 through the AIMES project ({\url{http://www.aimes.com.au}).  The speed data is reported at 15-minute intervals. We also divide this dataset by time into training (70\%), validation (10\%), and testing (20\%) subsets. This dataset has fewer sensors. However, it offers important insights to the model performance in a city CBD instead of highways in PEMS04 and PEMS08. We did not add an adjacency matrix to this dataset because none of the sensors are at adjacent intersections. 

Table~\ref{tab:error_na} summarizes the prediction errors for the models that can run without adjacency matrices. ARIMA, FC-LSTM,  WaveNet, and AGCRN do not use adjacent matrices. Their prediction errors do not change on PEMS04 and PEMS08. 

Without adjacent matrices, \model\ has slightly increased errors comparing with \model\ using adjacent matrices  (cf. Table~\ref{tab:error}). A similar observation is made for Graph WaveNet which also has a learned adjacency matrix. However, \model\ still 
outperforms all competitors. This confirms the effectiveness of LPGCN and the learned PMI matrix for spatial correlation modeling, and the robustness of \model\  without given adjacency matrices. On MEL, \model\  outperforms AGCRN by up to 1.5\% and 0.9\% in MAE and MAPE  (when predicting for 30 minutes ahead of time), respectively. This further confirms the robustness of \model\ over data from a city CBD.

\subsection{Effectiveness of the Temporal Correlation Module}\label{subsec:CNN_effect}

We further study the effectiveness our temporal correlation module (Section~\ref{subsec:temporal}). We compare \model\ with three variants: 
(1) \textbf{\model/spa}, which predicts with only the temporal correlation module, 
(2) \textbf{\model/tem}, which predicts with only LPGCN on the traffic observation at $t_p$ (and the prediction time $t_{p+i}$), and 
(3) \textbf{\model/att}, which replaces the temporal attention module with a fully connected layer to aggregate the multi-path CNN output. 

\begin{table}[ht]
\vspace{-2mm}
\centering
\setlength{\tabcolsep}{1.0mm}
    \caption{Prediction Errors with and without Temporal Correlation Modeling (RMSE results omitted for space limit)}
    \label{tab:error_temporal}
               \vspace{-2mm}
    {\begin{tabular}{l|l|rr|rr|rr}
        \toprule
        \multirow{2}{*}{Dataset} & \multirow{2}{*}{Model} & \multicolumn{2}{c|}{15 minutes} & \multicolumn{2}{c|}{30 minutes}& \multicolumn{2}{c}{60 minutes}\\ \cline{3-8}
         & & MAE  & MAPE & MAE  & MAPE  & MAE  & MAPE\\
        \midrule
        \midrule
\multirow{4}{*}{PEMS04}

& \model/spa& 1.06 & 1.77\%  & 1.24 &  2.16\% & 1.29 & 2.36\% \\ \cline{2-8} 
& \model/tem& 1.16 & 1.88\%& 1.28 & 2.35\% &1.46 & 2.74\% \\ \cline{2-8} 
&  \model/att  & 0.96 & 1.71\% & 1.19  & 2.21\% & 1.49 & 2.86\% \\ \cline{2-8}
& \textbf{\model}  & \textbf{0.96} & \textbf{1.71\%} & \textbf{1.12} & \textbf{2.05\%} & \textbf{1.21} & \textbf{2.14\%} \\ 
  \bottomrule
 
 \multirow{4}{*}{PEMS08}
& \model/spa& 1.34 & 2.49\%& 1.51 & 3.51\% &1.68 & 4.25\%\\ \cline{2-8} 
& \model/tem& 1.41 & 2.83\% & 1.63 & 3.84\% & 1.86 & 4.63\%\\ \cline{2-8} 
& \model/att & 1.35 & 2.57\% & 1.68 & 4.03\% & 2.16 & 5.19\%\\ \cline{2-8} 
& \textbf{\model}  & \textbf{1.22} & \textbf{2.41\%} & \textbf{1.43} & \textbf{3.43\%} & \textbf{1.61} & \textbf{4.03\%}\\ 
        \bottomrule

        \end{tabular}}
        \vspace{-2mm}
\end{table}

\textbf{Impact of the temporal correlation module.} 
From Table~\ref{tab:error_temporal}, we see that missing either the spatial (\model/spa) or the temporal (\model/tem) correlation module negatively impacts the prediction accuracy (comparing with the full \model).  \model/tem has higher prediction errors than  \model/spa, which highlights the importance of the temporal correlation module. 
By comparing \model/tem (no temporal correlation), \model/att (with multi-path CNN), and \model\ (with multi-path CNN and temporal attention), we see that, adding the multi-path CNN without temporal attention helps reduce the errors of predictions for 15 minutes ahead, but it does not necessarily help the predictions for further ahead. This is because the temporal correlation may vary at different time which needs to be modeled by the temporal attention module. This observation further confirms 
the importance of both our multi-path CNN and temporal attention modules. 

\textbf{Impact of $p$.} Next, we show the effectiveness of our temporal correlation module 
to learn the joint impact of historical traffic observations as $p$ grows from 1 to $q$, i.e., when more historical traffic observations are available. 
Fig.~\ref{fig:p_impact} shows the results in terms of MAE (MAPE and RMSE results are similar and are omitted), where 15, 30, and 60 denote predicting for 15, 30, and 60 minutes ahead, respectively. As $p$ increases, the prediction errors drop as expected. When $p$ increases from 1 to 12, a maximum drop of 17.1\% (from 1.46 to 1.21 when predicting for 60 minutes ahead on PEMS04) in MAE is observed. This confirms that our temporal correlation module can learn the joint impact when more historical traffic observations are available. 

\begin{figure}[ht]
\vspace{-5mm}
\begin{subfigure}{.520\linewidth}
  \includegraphics[width=1\linewidth]{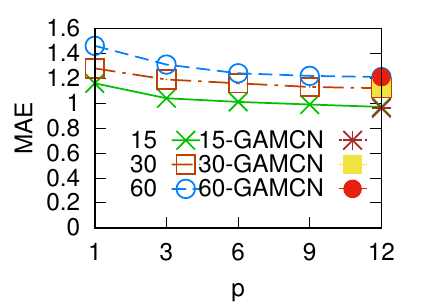}  
  \caption{PEMS04}
  \label{fig:p_pems04_mae}
\end{subfigure}
\hspace{-5mm}
\begin{subfigure}{.52\linewidth}
  \includegraphics[width=1\linewidth]{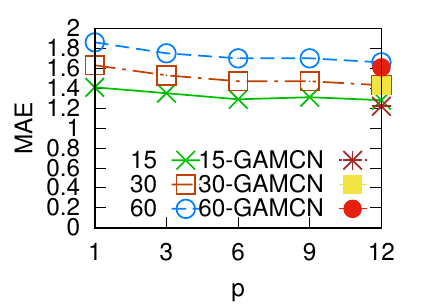}  
  \caption{PEMS08}
  \label{fig:p_pems08_mae}
\end{subfigure}
\vspace{-2mm}
\caption{Impact of $p$ on prediction errors}\label{fig:p_impact}
\vspace{-2mm}
\end{figure}

When $p < 12$, we do not have LPGCN output $\boldsymbol{Z_{t_i}}$ for every prediction time $t_{p+i}$. We thus have used $\boldsymbol{Z_{t_p}}$ for each $t_{p+i}$ in this set of experiments (even when $p=12$).  

When $p=12$, we also plot the result of using $\boldsymbol{Z_{t_i}}$ for each $t_{p+i}$  (i.e., the full \model), denoted by 15-\model, 30-\model, and 60-\model.  
We see that the model performance is very close using $\boldsymbol{Z_{t_i}}$ or $\boldsymbol{Z_{t_p}}$ (especially on PEMS04). This further confirms the effectiveness of \model\ in spatial correlation learning, even with just $\boldsymbol{X_{t_p}}$.

\subsection{Model Extendibility}\label{subsec:exp_gan}

As mentioned in Section~\ref{sec:lit_review}, recent models have used GANs for spatio-temporal data generation, e.g., TrafficGAN~\cite{zhang2021tits} for traffic prediction. They use a discriminator in addition to a generator (the prediction model) to  reduce the  distribution difference between predicted data and real data. 

In this subsection, we extend \model\ with GAN to improve the distribution similarity between its prediction and the ground truth, for better applicability in scenarios where not only the prediction errors but also the distribution similarity are important, e.g., when predicting for highly skewed traffic caused by large (e.g., sports) events. 

As Fig.~\ref{fig:gan} shows, we use \model\ as the generator $\mathcal{G}$ to generate predictions for the next $q$ time points. The discriminator $\mathcal{D}$ takes the traffic data (predicted or ground truth) of $q$ time points (i.e., $q$ matrices each in $\mathbb{R}^{|V| \times c}$) and aims to distinguish generated predictions from the ground truth. It  first uses two linear layers to map each matrix into an $\mathbb{R}^{|V|\times d}$ space ($d = 100$). Each mapped matrix then goes through LPGCN to learn the latent correlations. The LPGCN output is concatenated to form a $q|V|\times d$ matrix, which is fed into two linear layers  to generate the discriminator output. 

For model training, the generator \emph{minimizes} 
$\mathbb{E}_\mathbb{X}[\log(1-\mathcal{D}(\mathcal{G}({\boldsymbol{X_{t_1}}}, {\boldsymbol{X_{t_2}}}, \ldots, {\boldsymbol{X_{t_p}}}))]$, while 
the discriminator \emph{maximizes} $\mathbb{E}_\mathbb{X}[\log(\mathcal{D}({\boldsymbol{X_{t_p+1}}}, {\boldsymbol{X_{t_p+2}}}, \ldots, {\boldsymbol{X_{t_p+q}}}))] + \mathbb{E}_\mathbb{X}[\log(1-\mathcal{D}(\mathcal{G}({\boldsymbol{X_{t_1}}}, {\boldsymbol{X_{t_2}}}, \ldots, {\boldsymbol{X_{t_p}}}))]$. We train the generator for five epochs before every epoch of  discriminator training.  
We denote the resultant model as \emph{\model-GAN} and compare it with \model. We do not compare with TrafficGAN because its source code is unavailable (Source code of an earlier version of TrafficGAN~\cite{8970742} is available but it is based on Euclidean space which is inapplicable). Also, our aim is to show the applicability of \model\ with GANs, not to show its advantage over existing GAN-based models. 

\begin{figure}[H]
\vspace{-2mm}
\centering
\includegraphics[width=1\linewidth]{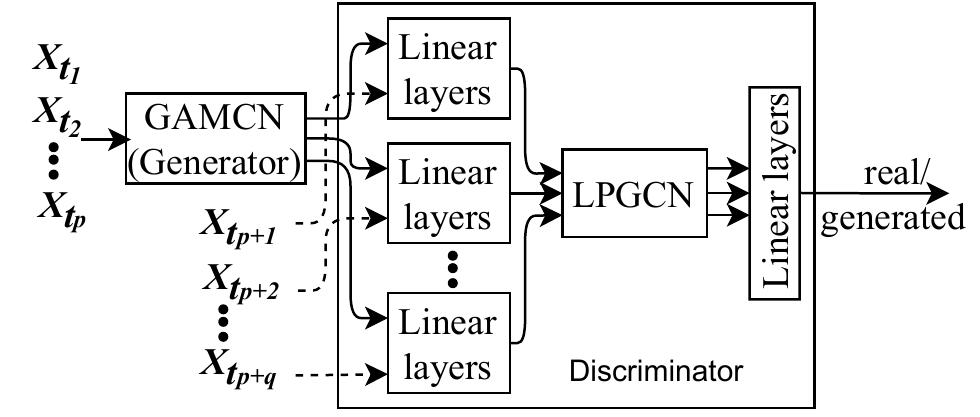} 
\caption{Model architecture of \model-GAN}
\label{fig:gan}
\vspace{-4mm}
\end{figure}

Table~\ref{tab:error_gan} presents the results on MAE and \emph{KL-divergence} (\textbf{KL}), which is a measurement of the similarity of two  probability distributions $P$ and $Q$ defined on the same probability space $\mathscr{X}$, $\mbox{KL}(P||Q) = \sum_{x\in\mathscr{X}}P(x)\log\frac{P(x)}{Q(x)}$. 

 \model-GAN has smaller KL values than \model\ on both datasets in all different prediction time settings, confirming the effectiveness of a GAN-based model in reducing the distribution difference between the prediction and the ground truth. However, \model-GAN also has larger MAE, which is less preferable. To further reduce MAE, we adapt the loss function of the generator to incorporate the MSE loss $\mathcal{L}$ (Equation~\ref{eq:mse_loss}). The generator now minimizes $\lambda\mathcal{L} + \mathbb{E}_\mathbb{X}[\log(1-\mathcal{D}(\mathcal{G}({\boldsymbol{X}}_{t_1}, {\boldsymbol{X}}_{t_2}, \ldots, {\boldsymbol{X}}_{t_p}))]$, where $\lambda = 0.01$ is a hyper-parameter.  We denote this resultant model by \model-GAN$^*$. This model yields MAE as small as those of \model\ and KL as small as those of \model-GAN in most cases. This is desirable,  and it confirms the applicability of our model in producing predictions that not only have small errors overall but also follow the real data distribution. 

\begin{table}[ht]
\centering
\setlength{\tabcolsep}{1.45mm}
    \caption{Prediction Errors of \model\ with GAN}
    \label{tab:error_gan}
               \vspace{-2mm}
    {\begin{tabular}{l|l|rr|rr|rr}
        \toprule
        \multirow{2}{*}{Dataset} & \multirow{2}{*}{Model} & \multicolumn{2}{c|}{15 minutes} & \multicolumn{2}{c|}{30 minutes}& \multicolumn{2}{c}{60 minutes}\\ \cline{3-8}
         & & MAE  & KL & MAE  & KL  & MAE  & KL\\
        \midrule
        \midrule
\multirow{4}{*}{PEMS04}

& \model& \textbf{0.96} & 0.03  & \textbf{1.12} &  0.05 & \textbf{1.21} & 0.8 \\ \cline{2-8} 
& \model-GAN& 1.01 & \textbf{0.02}& 1.15 & \textbf{0.04} & 1.27 & \textbf{0.6}\\ \cline{2-8} 
& \model-GAN$^*$ & 0.97 & \textbf{0.02} & 1.13 & \textbf{0.04} & \textbf{1.21} & 0.7 \\ 
  \bottomrule
 
 \multirow{4}{*}{PEMS08}
& \model& \textbf{1.22} & 0.07& \textbf{1.43} & 0.12 & \textbf{1.61} & 0.15 \\ \cline{2-8} 
& \model-GAN& 1.29 & \textbf{0.06}& 1.47 & \textbf{0.11} & 1.65 & \textbf{0.12}\\ \cline{2-8} 
& \model-GAN$^*$ & \textbf{1.22} & \textbf{0.06} & 1.44 & \textbf{0.11} & \textbf{1.61}&0.13 \\ 
        \bottomrule

        \end{tabular}}
        \vspace{-4mm}
\end{table}

\section{Conclusions and Future Work}\label{sec:conclusion}
We proposed a model named \model\ for predicting traffic conditions on road networks in the short-term future. \model\ consists of an improved GCN component to learn the spatial correlations of traffic conditions and an attentive multi-path  CNN component to learn the temporal correlations of traffic conditions. Experiments on real-world traffic datasets show that \model\ has substantially lower (up to 18.9\%) traffic prediction errors than state-of-the-art models, while its time efficiency is higher (by up to 23.4\%),  making our model highly practical for real-world applications. 

\model\  training can be accelerated using TensorFlow distributed training APIs (\url{https://www.tensorflow.org/guide/distributed_training}) with minimal code changes, 
which distribute  training samples across computation units (e.g., GPUs).  When the road network of interest is large, or there is a large number of traffic sensors, 
a single computation node may not have sufficient memory to accommodate a full input traffic observation graph.  
We  plan to parallelize \model\  for such scenarios, to further enhance the model applicability. The main bottleneck lies in LPGCN, since graph neural networks are  expensive to train in terms of both time and memory space costs. Existing distributed graph neural network systems such as AliGraph~\cite{DBLP:journals/pvldb/ZhuZYLZALZ19} and DistDGL~\cite{DBLP:journals/corr/abs-2010-05337} do not provide APIs to learn and update  the graph adjacency matrices during model training, which is required by LPGCN. It would be interesting to extend such systems to support LPGCN and hence \model.


%



\vspace{-3mm}
\ifCLASSOPTIONcompsoc
 \section*{Acknowledgments}
\else
   regular IEEE prefers the singular form
\section*{Acknowledgment}
\fi
This work is partially supported by the FEIT Platform Interdisciplinary Grant (2020) of The University of Melbourne.
\vspace{-5mm}

\ifCLASSOPTIONcaptionsoff
  \newpage
\fi



\bibliographystyle{IEEEtran}
\bibliography{ref}

%

%

\vspace{-12mm}

\begin{IEEEbiography}[{\includegraphics[width=1in,height=1.25in,clip]{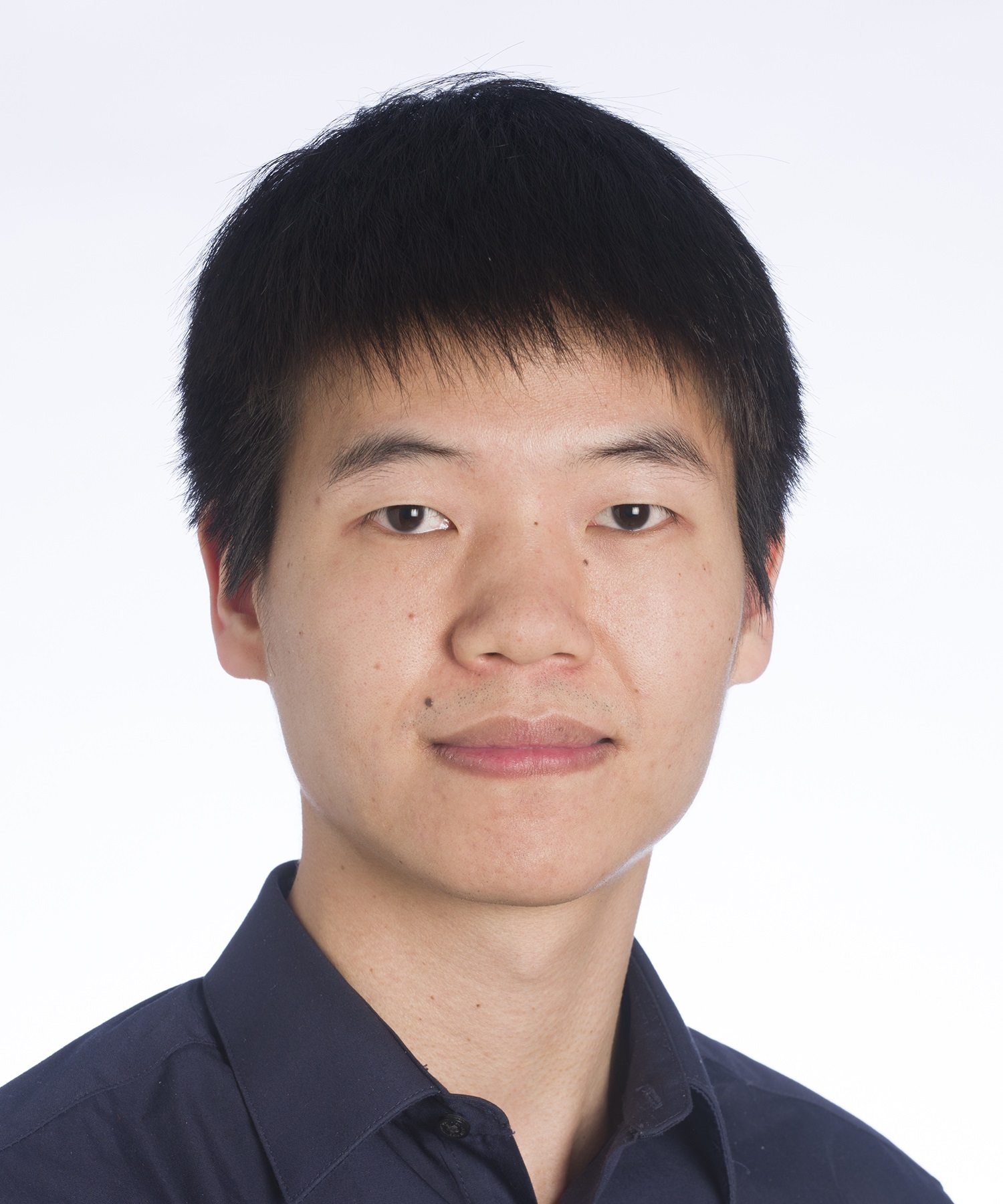}}]{Jianzhong Qi}
is a Senior Lecturer in the School of Computing and Information Systems, The University of Melbourne. 
He received his Ph.D. degree from The University of Melbourne. 
His research interests include machine learning and data management and analytics, with a focus on spatial, temporal, and textual data.
\end{IEEEbiography}

\vspace{-12mm}
\begin{IEEEbiography}[{\includegraphics[width=1in,height=1.25in,clip]{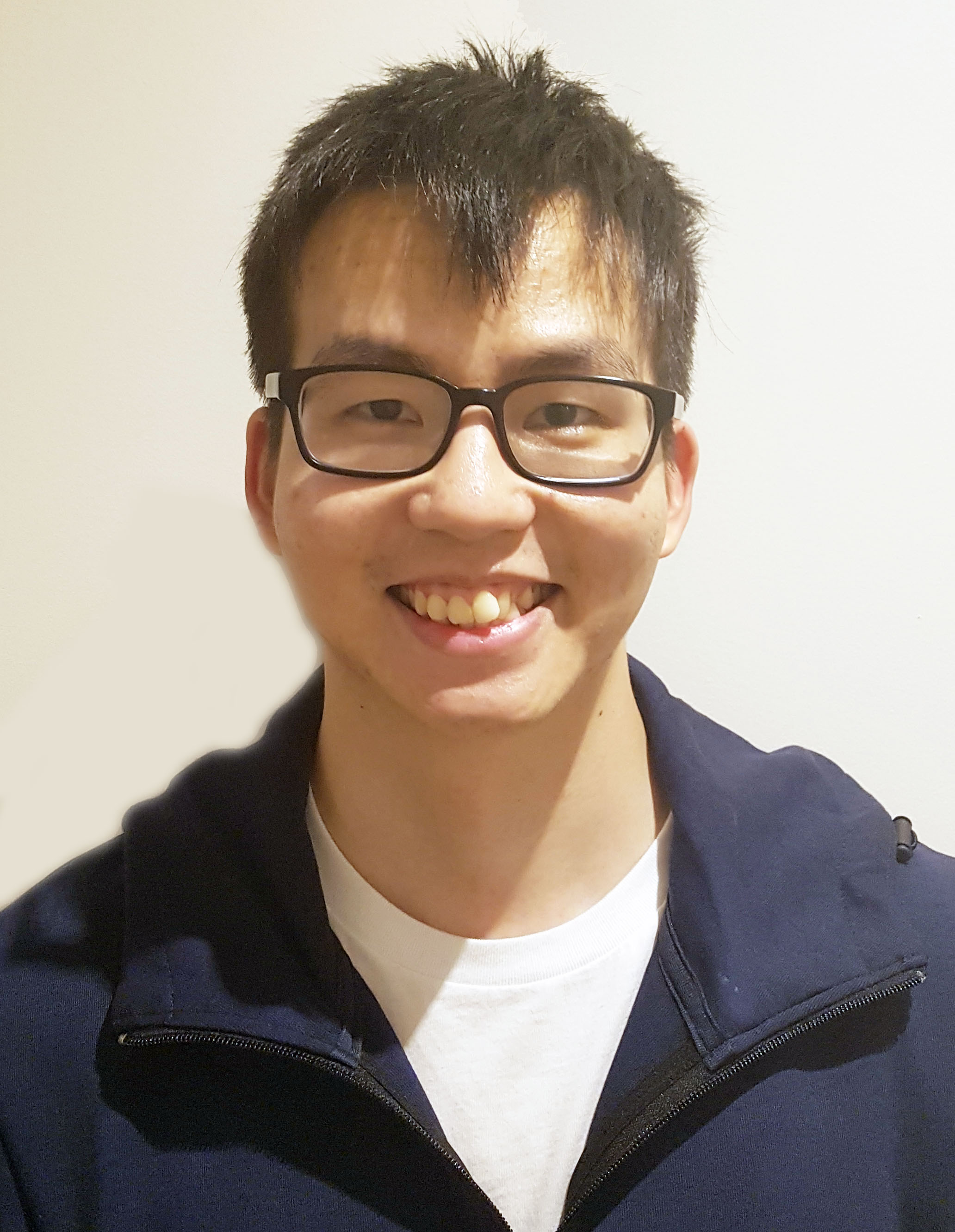}}]{Zhuowei Zhao} is a Ph.D. student in the School of Computing and Information Systems, The University of Melbourne.  He received his MPhil degree  from The University of Melbourne in 2020. His research focuses on machine learning for graph data.
\end{IEEEbiography}


\vspace{-12mm}

\begin{IEEEbiography}[{\includegraphics[width=1in,height=1.25in,clip]{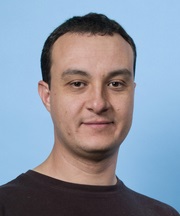}}]{Egemen Tanin}
is a Professor in the School of Computing and Information Systems, The University of Melbourne. 
He received his Ph.D. degree from the University of Maryland at College Park. 
His research interests include spatial databases and mobile data management. 
\end{IEEEbiography}

\vspace{-12mm}

\begin{IEEEbiography}[{\includegraphics[width=1in,height=1.25in,clip]{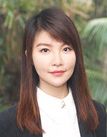}}]{Tingru Cui}
is a Senior Lecturer in the School of Computing and Information Systems, The University of Melbourne. 
She received her Ph.D. degree from the National University of Singapore. 
Her research interests are interdisciplinary in nature, exploring the human-AI interaction, business and learning analytics, social media, crowdsourcing, and digital innovation. 
\end{IEEEbiography}

\vspace{-12mm}

\begin{IEEEbiography}[{\includegraphics[width=1in,height=1.25in,clip]{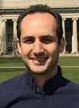}}]{Neema Nassir}
is a Lecturer in the Department of Infrastructure Engineering, The University of Melbourne. 
He received his Ph.D. degree from the University of Arizona. 
His research focuses on new methods to simulate, model, design and manage public transport, shared-mobility, and connected multimodal systems. 
\end{IEEEbiography}

\vspace{-12mm}

\begin{IEEEbiography}[{\includegraphics[width=1in,height=1.25in,clip]{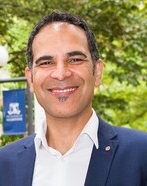}}]{Majid Sarvi}
is a Professor in the Department of Infrastructure Engineering, The University of Melbourne. 
His fields of research cover a range of topics, including: connected multi-modal transport network, crowd dynamic modeling and simulation, and network vulnerability assessment and optimization. 
\end{IEEEbiography}




\end{document}